\documentclass[10pt,journal,compsoc]{IEEEtran}

\ifCLASSOPTIONcompsoc
  \usepackage[nocompress]{cite}
\else
  \usepackage{cite}
\fi

\usepackage{amsmath,amssymb,amsthm,amsfonts}
\usepackage{graphicx}
\usepackage{multirow}
\usepackage{booktabs}
\usepackage{algorithm,algorithmic}
\usepackage{enumitem}
\usepackage[caption=false,font=footnotesize]{subfig}
\usepackage{bm}
\usepackage{dsfont}
\usepackage{tikz}
\usepackage{ragged2e}
\usepackage{makecell}
\usepackage[normalem]{ulem}
\usepackage{hyperref}
\usepackage{bbm}
\usepackage[utf8]{inputenc}
\usepackage[english]{babel}
\usepackage{epstopdf}
\usepackage{array}
\usepackage{svg}
\usepackage{pifont}
\usepackage{amsmath}

\usepackage[most]{tcolorbox}  
\usepackage{colortbl}

\definecolor{darkblue}{rgb}{0, 0, 0.5}
\definecolor{myblue}{rgb}{0.0, 0.0, 0.6}
\hypersetup{colorlinks=true, citecolor=darkblue, linkcolor=darkblue, urlcolor=darkblue}

\begin{document}

\title{Efficient Inference for \\ Large Reasoning Models: A Survey}

\author{Yue~Liu,
        Jiaying~Wu,
        Yufei~He, 
        Ruihan~Gong, Jun~Xia, Liang~Li,\\
        Hongcheng~Gao,
        Hongyu~Chen,
        Baolong~Bi,
        Jiaheng~Zhang,
        Zhiqi~Huang,\\
        Bryan~Hooi,
        ~Stan~Z.~Li,~\IEEEmembership{Fellow,~IEEE}
        and~Keqin~Li,~\IEEEmembership{Fellow,~IEEE}

\IEEEcompsocitemizethanks{
\IEEEcompsocthanksitem Yue Liu, Jiaying Wu, Yufei He, and R. Gong have equal contributions.
\IEEEcompsocthanksitem Yue Liu, Jiaying Wu, Yufei He, Ruihan Gong, Liang Li, Jiaheng Zhang, and Bryan Hooi are with National University of Singapore.
\IEEEcompsocthanksitem Jun Xia is with HKUST (Guangzhou).
\IEEEcompsocthanksitem Hongcheng Gao, Hongyu Chen, and Baolong Bi are with UCAS.
\IEEEcompsocthanksitem Zhiqi Huang is with Moonshot AI.
\IEEEcompsocthanksitem Stan Z. Li is with Westlake University.
\IEEEcompsocthanksitem Keqin Li is with State University of New York.
}
\thanks{Manuscript received 13th August, 2025.}}

\IEEEtitleabstractindextext{%
\justifying
\begin{abstract}
Large Reasoning Models (LRMs) significantly improve the reasoning ability of Large Language Models (LLMs) by learning to reason, exhibiting promising performance in solving complex tasks. 
However, their deliberative reasoning process leads to inefficiencies in token usage, memory consumption, and inference time. 
Thus, this survey provides a review of efficient inference methods designed specifically for LRMs, focusing on mitigating token inefficiency while preserving the reasoning quality. 
The overview structure of this paper is shown in Figure~\ref{fig:paper_structure}. 
First, we introduce a taxonomy to group the recent methods into two main categories: (a) explicit compact Chain-of-Thought (CoT), which reduces tokens while keeping the explicit reasoning structure, and (b) implicit latent CoT, which encodes reasoning steps within hidden representations instead of explicit tokens. Meanwhile, we discuss their strengths and weaknesses. Then, we conduct empirical analyses on existing methods from reasoning scenarios, object functions, and performance \& efficiency aspects. Besides, we present open challenges in this field, including human-centric controllable reasoning, trade-off between interpretability and efficiency of reasoning, ensuring the safety of efficient reasoning, and broader applications of efficient reasoning. In addition, we highlight key insights for enhancing LRMs' inference efficiency via techniques such as model merging, new architectures, and agent routers. We hope this work serves as a valuable guide, helping researchers overcome challenges in this vibrant field. A collection of efficient reasoning methods for LRMs (papers and codes) is provided at this link: \url{https://github.com/yueliu1999/Awesome-Efficient-Inference-for-LRMs}.
\end{abstract}

}

\maketitle

\section{Introduction}
\IEEEPARstart{L}{arge} Language Models (LLMs), which are trained to provide quick and intuitive responses, have exhibited great success in various complex fast-thinking applications like ChatBot \cite{ChatGPT}. Slow-thinking scenarios like math problem-solving \cite{AIME} or research \cite{deep_research} increasingly require the models to conduct advanced analytical and deliberative reasoning before providing final responses. To tackle these challenges, Large Reasoning Models (LRMs) such as OpenAI o1/o3 \cite{jaech2024openai,openai2025systemcard}, DeepSeek R1 \cite{guo2025deepseek}, and Kimi k1.5 \cite{team2025kimi} are developed by guiding the model to learn to effectively reason.

\begin{figure*}[!t]
\centering
\label{fig:paper_structure}
\includegraphics[width=0.9\textwidth]{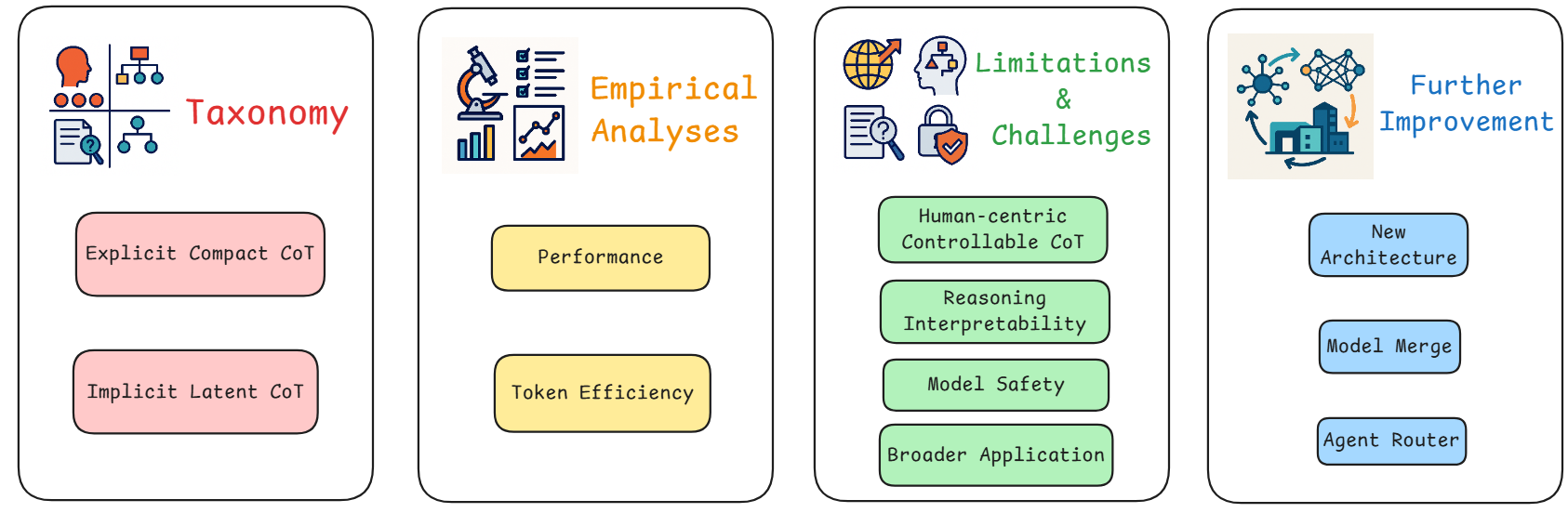}
\caption{\textbf{Overview of this Survey}. It mainly consists of four parts: taxonomy, empirical analyses, limitations \& challenges, \& further improvement.}
\end{figure*}

Although effective, the intermediate reasoning process of LRMs is highly resource-intensive, learning to three challenges: (1) significant token consumption, (2) high memory overhead, and (3) increased inference time. Bottlenecks in the safety fine-tuning of vision-language models, as discussed in \cite{ding2025rethinking}, can severely impact their deployment in critical applications, where model reliability and trustworthiness are paramount.
These problems not only increase the inference cost of the service companies but also degrade the experience of the users. Therefore, efficient inference for LRMs has become an urgent and crucial direction. Since thinking tokens are treated like regular output tokens without cost differentiation, previous efforts in inference efficiency of regular LLMs, e.g., model compression \cite{wang2024model}, efficient model design \cite{llada}, and system-level optimization \cite{deepseek_v3}, can alleviate problems (2) and (3). These methods are comprehensively studied \cite{survey_1} and not specially designed for LRMs. Therefore, this survey focuses on the challenge (1): token inefficiency, as shown in Figure~\ref{fig:paper_structure}.

\begin{figure*}[htbp]
\centering
\includegraphics[width=0.9\textwidth]{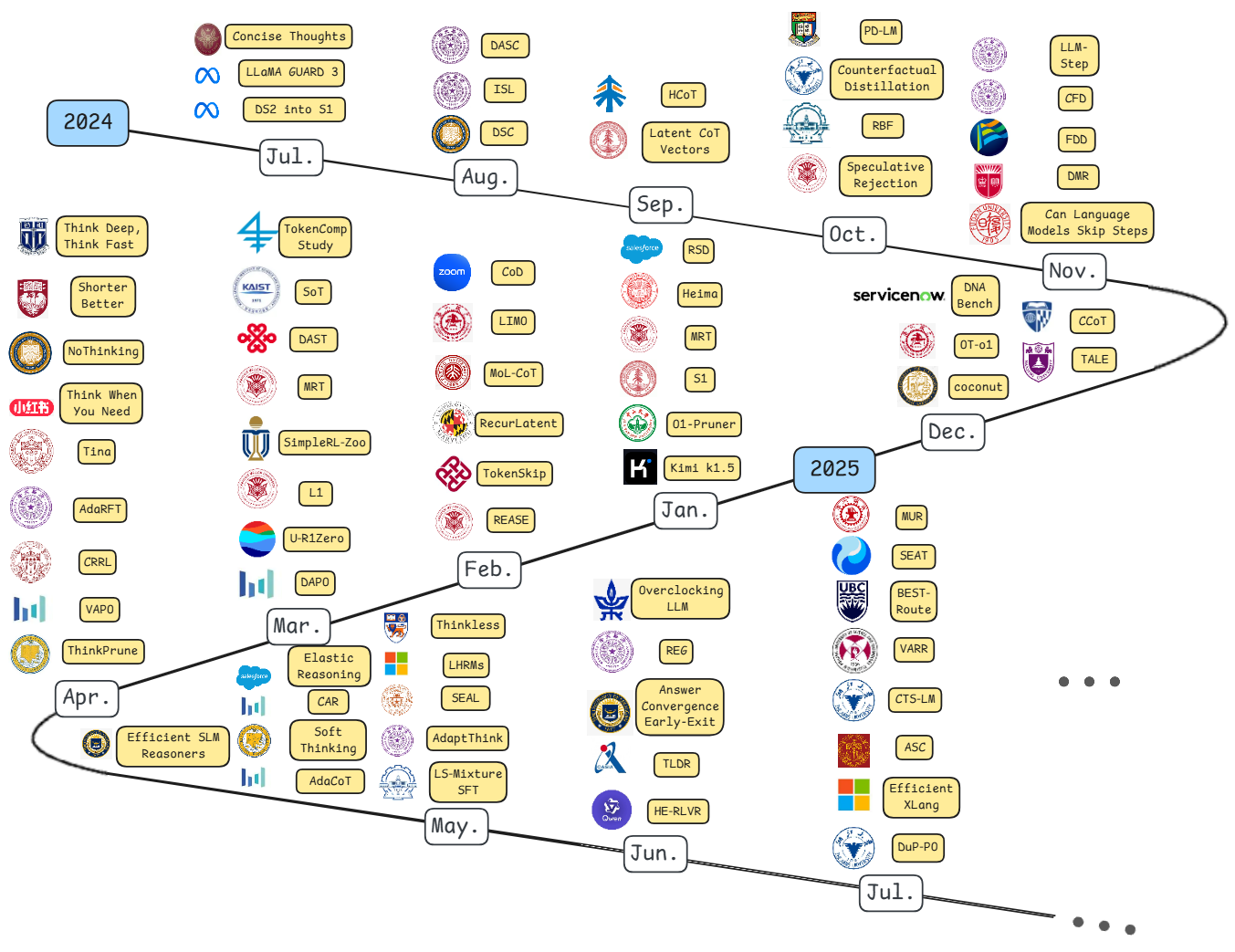}
\caption{\textbf{Chronological Milestones of Efficient Inference for Large Reasoning Models.} The time range is mainly from July 2024 to July 2025. }
\label{fig:timeline_high_cited}
\end{figure*}

To this end, we conduct a comprehensive survey of recent efficient inference methods designed specifically for LRMs, aiming at improving thinking token efficiency while preserving reasoning quality. Concretely, we first illustrate the research landscape over time as shown in Figure~\ref{fig:timeline_high_cited}, which presents a chronological overview of selected highly-cited papers on efficient inference for LRMs from July 2024 to July 2025.
This timeline highlights representative works that have had a notable impact in the community, rather than providing an exhaustive or complete list.
It serves to contextualize the subsequent discussion by showing when key contributions appeared over the past year.

\begin{figure*}[!t]
\centering
\label{fig:overview}
\includegraphics[width=0.9\textwidth]{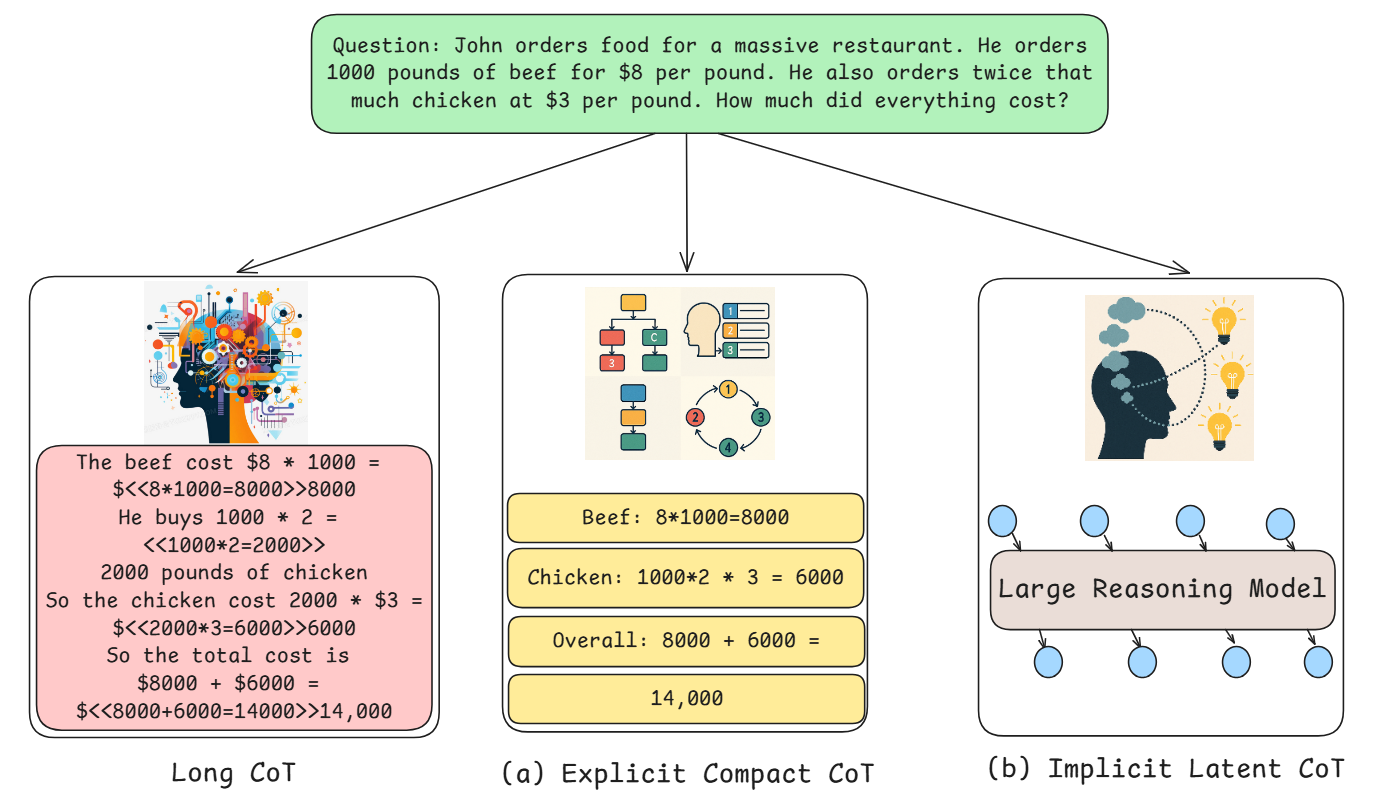}
\caption{\textbf{Taxonomy of Efficient Inference for Large Reasoning Models.} The large reasoning model typically outputs long CoT (left sub-figure). The recent efficient inference methods for large reasoning models are mainly classify into (a) explicit compact CoT and (b) implicit latent CoT.}
\end{figure*}

Subsequently, we present a hierarchical taxonomy that categorizes recent approaches into two classes. As shown in Figure \ref{fig:overview}, it contains (a) the explicit compact CoT, which reduces the number of thinking tokens while maintaining explicit reasoning structure, and (b) the implicit latent CoT, which encodes reasoning steps within hidden representations instead of explicit tokens. In addition, for the explicit compact CoT, we further summarize three sub-categories: (a.1) CoT compression, (a.2) CoT preference optimization, and (a.3) reward-based CoT conciseness. We analyze the characteristics and discuss their strengths and weaknesses from the aspects of reasoning quality and efficiency.

Moreover, we conduct a comprehensive empirical study on the existing methods from the perspectives of reasoning scenarios, object functions, and performance \& efficiency aspects. Besides, we identify four open challenges regarding the inference efficiency of LRMs, including human-centric controllable reasoning, the trade-off between efficiency and interoperability of reasoning, ensuring the safety of efficient reasoning, and broader applications of efficient LRMs beyond math and code. Lastly, we highlight potential techniques for further improving current methods, including model merging, new architectures, and agent routers.

We hope that this survey helps researchers and engineers further improve efficient inference for LRMs. The main contributions of this paper are summarized as follows.

\begin{enumerate}[label=\textbullet, leftmargin=0.4cm, itemsep=0.2em, parsep=0.2em, topsep=0.em]  
    \item We conduct a comprehensive paper review of current methods of efficient inference for LRMs with a hierarchical taxonomy and strength \& weakness discussion. 
    
    \item We empirically study recent methods from reasoning scenarios, object functions, and performance \& efficiency.

    \item We summarize four challenges in this domain from user control, interpretability, safety, and application aspects.

    \item We highlight technical insights in further improvement of existing methods from the perspectives of model merging, non-autoregressive architectures, and agent routers
\end{enumerate}

\section{Background}
This section first introduces the background of large reasoning models and then highlights the efficiency challenges in the inference phase of large reasoning models. 

\subsection{Large Reasoning Model}
Large Reasoning Models (LRMs) extend the capabilities of Large Language Models (LLMs) by incorporating explicit intermediate tokens that represent reasoning processes, enabling more structured logical reasoning and effective complex problem-solving.
LRMs mimic the way humans approach complex problems by first \textbf{thinking} before providing an \textbf{answer}.
When faced with a difficult question, they do not immediately respond with an answer; instead, they analyze the problem, break it down into smaller steps, explore different solution paths, and verify their reasoning before arriving at a conclusion.
This human-like reasoning process of LRMs can also be examined through a cognitive framework, as discussed in \cite{hu2025unveiling}, which provides insights into the underlying mechanisms shaping model reasoning behaviors.
The o1 series~\cite{jaech2024openai} from OpenAI, released in late 2024, marked a significant breakthrough in AI reasoning capabilities, which integrates reinforcement learning and "Chain of-Thought" prompting~\cite{wei2022chain} techniques. Following this, OpenAI released o3~\cite{openai2025systemcard}, an upgraded version of o1, allowing it to achieve PhD-level performance in mathematics, science, and programming.
Notable DeepSeek's R1~\cite{guo2025deepseek} stands out for being fully open-sourced, with transparent and detailed thinking process tokens, which sets it apart from other proprietary LRMs like o1/o3, where the internal reasoning steps are less accessible.
However, since LRMs need to generate numerous intermediate thinking tokens over before arriving at final answers, they are significantly less efficient and more expensive compared to regular LLMs. 
This added complexity in processing demands significantly more computational resources and time.

\subsection{Efficiency Challenge in LRM Inference}

A key driver of LRMs' remarkable reasoning capabilities is the scaling of inference-time compute, which enables complex reasoning through long CoTs \cite{chen2025towards,guo2025deepseek,jaech2024openai,muennighoff2025s1,liu2025bag}. Compared to standard short CoTs \cite{wei2022chain}, which are often shallow, heuristic-driven, and less generalizable \cite{sprague2025to}, long CoTs empower LRMs to tackle complex tasks such as advanced mathematics \cite{xu2025redstar} and medical question answering \cite{huang2025o1}. However, this shift has also introduced the phenomenon of overthinking, where LRMs consume excessive inference tokens and reasoning steps even for simple problems, yielding only marginal performance improvements \cite{ma2024step,over_thinking,wu2025more}. In real-world applications such as software engineering agents, overthinking has been found to negatively correlate with issue resolution rates \cite{cuadron2025danger}. Moreover, LRMs' reliance on inference-time scaling exposes them to overthinking attacks, where adversarial actors inject benign yet computationally intensive decoy problems (e.g., Sudoku puzzles) into the context for retrieval-augmented question answering, triggering substantial computational overhead \cite{kumar2025overthink}.

Toward practical and scalable real-world deployment, optimizing the token efficiency of LRMs without compromising overall effectiveness remains an underexplored challenge. This paper presents a comprehensive and systematic investigation into recent advances in token-efficient LRMs, examining their underlying approaches, empirical effectiveness, and implications for future research.

\section{Landscape of Efficient Reasoning}

This section surveys the current landscape of research on token-efficient LRM inference, which can be broadly categorized into two approaches: (1) \textbf{explicit compact CoT}, where explicit instructions, rewards, or budget constraints are introduced to encourage shorter reasoning chains over long CoTs (Section~\ref{sec:explicit-cot}); and (2) \textbf{implicit latent CoT}, which compresses explicit long CoTs into compact, continuous reasoning states (Section~\ref{sec:implicit-cot}). The taxonomy of recent efficient inference methods is shown in Table \ref{tab:taxonomy_3_1}, \ref{tab:taxonomy_3_2} and \ref{tab:taxonomy_4}, providing a detailed breakdown of both explicit compact CoT and implicit latent CoT methods in terms of their strategies, training regimes, models, and application domains.

\begin{table*}[!t]
\centering
\setlength{\tabcolsep}{3pt}
\caption{\label{tab:taxonomy_3_1}\textbf{Taxonomy of Explicit Compact CoT Methods (Part I).} The criteria mainly contain training, strategy, model, and application. }
\resizebox{\linewidth}{!}{
\begin{tabular}{lccccc}
\toprule[1.1pt]
\textbf{Types} & \textbf{Methods}  & \textbf{Training} & \textbf{Strategy} & \textbf{Model} & \textbf{Application} \\
\midrule
\multirow{50}{*}{\shortstack[l]{Explicit \\ Compact \\CoT}} 
& SoT \cite{aytes2025sketch} & \ding{55} & Prompt &  Qwen-2.5-7B/14B/32B & Math, Commonsense, Logic, Scientific, Medical\\
 & Constrained-CoT \cite{nayab2024concise} & \ding{55} & Prompt & LLaMA-2-70B, Falcon-40B & Math \\ 
 & CoD \cite{xu2025chain} & \ding{55} & Prompt & GPT-4o, Claude 3.5 Sonnet & Math, Commonsense, Symbolic Reasoning \\ 
& TALE-EP \cite{TALE} & \ding{55} & Prompt & LLaMA-3.1-8B-Instruct & Math \\ 
 & Meta-Reasoner \cite{meta_reasoner} & \ding{55} & Prompt & GPT-4o, GPT-4o-mini, Gemini-Exp-1206 & Math, Scientific \\ 
 & TS \cite{zhang2025making} & \ding{55} & Intervention & Qwen-2.5-7B/14B/32B & Math\\
 & Fractured Sampling \cite{liao2025fractured} & \ding{55} & Inference-time Scaling & DeepSeek-R1/Qwen-1.5B/7B/14B & Math, Scientific, Logic\\
 & RPC \cite{song2025reasoning} & \ding{55} & KV Cache Compression & QwQ-32B/DeepSeek-R1-Distill-Qwen-7B & Math, Code, Instruction \\
 & ThinkLess \cite{li2025thinkless} & \ding{55} & Prompt & Qwen-2.5-7B/14B, LLaMA3.1-8B & Math, Commonsense, Logic, Scientific\\
 & PLAN-AND-BUDGET \cite{lin2025plan} & \ding{55} & Prompt & \makecell{DeepSeek-R1-Distill-Qwen-32B, \\QwQ-32B, \\DeepSeek-R1-Distill-LLaMA-70B, \\OpenAI o4-mini} & Math, Instruction, Planning\\
 & TrimR \cite{lin2025trimr} & \ding{55} & Prompt & \makecell{Pangu Pro MoE, \\Pangu-R-38B, \\QwQ-32B, \\DeepSeek-R1-Distill-Qwen-32B} & Math, Scientific\\
& SOLAR \cite{li2025solar} & \checkmark & SFT & Qwen2VL-7B-Instruct & Math\\
& C3oT \cite{kang2024c3ot} & \checkmark & SFT & LLaMA-2-Chat
-7B \& -13B & Math, Commonsense \\
& TokenSkip \cite{xia2025tokenskip} & \checkmark & SFT & LLaMA-3.1-8B-Instruct,  Qwen2.5-
14B-Instruct & Math\\
& InftyThink \cite{yan2025inftythink} & \checkmark & SFT & \makecell{Qwen2.5-14B/32B, Qwen2.5-Math-1.5B/7B,\\ LLaMA-3.1-8B} & \makecell{Math,\\ Scientific} \\
& LightThinker \cite{zhang2025lightthinker} & \checkmark & SFT & \makecell{DeepSeek-R1-Distill-Qwen-7B,\\ DeepSeek-R1-Distill-LLaMA-8B} & \makecell{Language Understanding,\\ Math, Scientific, Commonsense,\\ Logic} \\
& CoT-Valve \cite{ma2025cot} & \checkmark & SFT & \makecell{QwQ-32B-Preview,\\ DeepSeek-R1-Distill-LLaMA-8B,\\ LLaMA-3.1-8B, LLaMA-3.2-1B,\\ Qwen32B-Instruct} & Math \\
& Distill System 2 \cite{yu2024distilling}  & \checkmark  & SFT &  LLaMA-2-70B-chat & Math, Commonsense, Coin Flip \\
& SF \cite{munkhbat2025self} & \checkmark  & SFT & \makecell{LLaMA-3.2-3B,  Gemma2-2B ,  Qwen2.5-3B , \\ Qwen2.5-Math-1.5B,  DeepSeekMath-7B}  & Math \\
   & Skip Steps \cite{liu2024can} & \checkmark & SFT & LLaMA2-7b, Phi-3-mini & Math, Logic\\

& DAST \cite{DAST} & \checkmark & SimPO  & \makecell{DS-R1-Distill-Qwen-7B, \\DS-R1-Distill-Qwen-32B}& Math\\
& TALE-PT \cite{TALE} & \checkmark & SFT, DPO & LLaMA-3.1-8B-Instruct & Math \\ 
& Kimi k1.5 \cite{team2025kimi} & \checkmark & RL & Kimi k1.5 & Multimodal Understanding, Math, Code\\
& O1-Pruner \cite{luo2025o1} & \checkmark & RL & Marco-o1-tB, QwQ-32B & Math\\
& MRT \cite{qu2025optimizing} & \checkmark & RL & DeepSeek-R1-Distill-Qwen-32B & Math\\
 &ERL \cite{arora2025training}  & \checkmark & RL & DS-R1-Distill-Qwen-1.5B, DS-R1-Distill-Qwen-7B & Math\\
& Claude 3.7 \cite{claud3_7} & \checkmark & RL & Unknown & Math, Code, Agent \\ 
 & L1 \cite{l1} & \checkmark & RL & Qwen-Distilled-R1-1.5B & Language Understanding, Logic, Math \\ 
 & SPIRIT \cite{cui2025stepwise} & \checkmark & RL & LLaMA3-8B-Instruct, Qwen2.5-
7B-Instruct  & Math \\
& IBPO \cite{yu2025think} & \checkmark & RL  & LLaMA-3.1-8B & Math\\
& LS-Mixture SFT \cite{yu2025long} & \checkmark & SFT & Qwen2.5-32B-Instruct & Math\\
& ConCISE \cite{qiao2025concise} & \checkmark & SFT, SimPO & DeepSeek-R1-Distill-Qwen-7B/1.5B & Math, Reasoning\\
& Elastic Reasoning \cite{xu2025scalable} & \checkmark & RL & E1-Math-1.5B/E1-Code-14B & Math, Code\\
& S-GRPO \cite{dai2025sgrpo} & \checkmark & RL & Qwen3-8B/14B, DeepSeek-R1-Distill-Qwen-7B/14B & Math, Scientific \\
& TLDR \cite{zhang2025making} & \checkmark & RL & Qwen-2.5-7B/14B/32B & Math\\
& Adaptive GoGI-Skip \cite{zhuang2025accelerating} & \checkmark & SFT &  Gemma3-1B/4B/12B, Qwen2.5-3B/7B & Math\\
\bottomrule[1.1pt]
\end{tabular}}
\end{table*}

\subsection{Explicit Compact CoT}
\label{sec:explicit-cot}

Recent research has focused on developing effective methods to create more compact reasoning paths while preserving high accuracy through various techniques, including (1) \textbf{CoT compression}, (2) \textbf{fine-tuning for compact reasoning}, and (3) \textbf{reward-based incentivization}.
To address the limitations of current reasoning models, Sherlock, as proposed by Ding and Zhang \cite{ding2025sherlock}, introduces a self-correcting mechanism that enhances the accuracy of vision-language models during inference.


\subsubsection{CoT Compression}

\begin{figure}[ht]
\centering
\small
\begin{minipage}{1.0\linewidth}
\centerline{\includegraphics[width=1\textwidth]{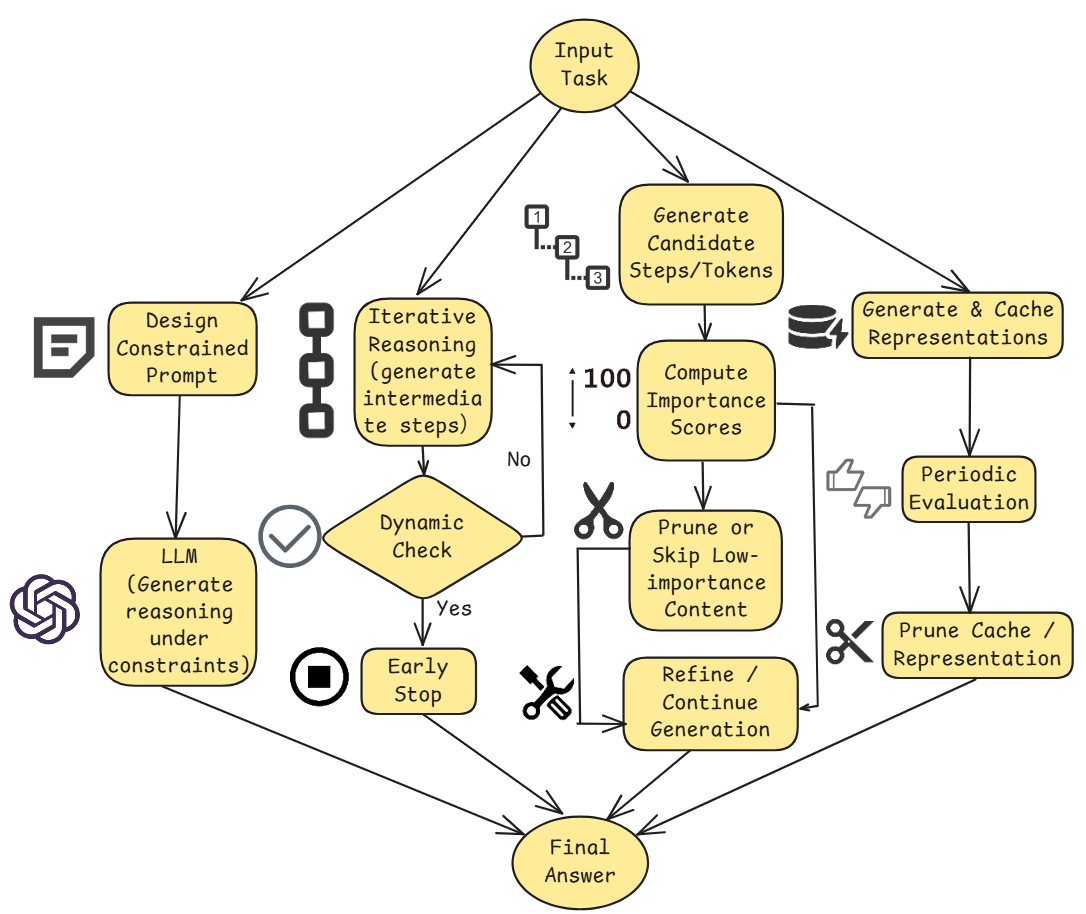}}
\end{minipage}
\caption{\textbf{Flowchart of CoT Compression Methods.} Each column represents one distinct kind of approach for compressing the CoT reasoning process, highlighting the key steps of each method.}
\label{fig:cot_compression_flow}
\end{figure}

Succinct CoT representations effectively streamline inference while preserving solution quality. The diagram in Figure~\ref{fig:cot_compression_flow} highlights the core steps of each approach, facilitating clear comparison and comprehensive understanding of the different techniques employed for CoT compression.

Several methods directly constrain the reasoning process to essential steps: 
Constrained-CoT \cite{nayab2024concise} and CoD \cite{xu2025chain} confine intermediate reasoning to essential steps, ensuring consistent brevity without losing critical information.  
Sketch-of-Thought (SoT) \cite{aytes2025sketch} uses a smaller ``router'' model to prompt the main LLM to generate sketches of reasoning, offering a concise yet cognitively inspired and structured overview.  
Fractured Sampling~\cite{liao2025fractured} interpolates between full CoT and direct-answer generation by recombining partial reasoning traces, enhancing accuracy-cost efficiency significantly without requiring model retraining.  
InftyThink \cite{yan2025inftythink} decomposes complex tasks into bounded-length segments, creating context-rich intermediate summaries at each step.

Some approaches dynamically adapt compression at inference time:
CoThink~\cite{fan2025cothink} uses an instruct model to guide concise solution outlines, improving token efficiency without accuracy loss.
ConCISE~\cite{qiao2025concise} applies confidence-guided early stopping to compress reasoning chains, reducing output length while maintaining accuracy.
ThinkLess~\cite{li2025thinkless} introduces early terminators and lightweight output regulation, reducing token overhead without extra training.
NoWait~\cite{wang2025wait} eliminates filler tokens using a training-free suppression method, producing concise outputs without affecting accuracy.
SReF~\cite{liu2025efficient} suppresses self-affirming reflections, shortening outputs without degrading accuracy across benchmarks.
Adaptive GoGI-Skip~\cite{zhuang2025accelerating} combines goal-gradient importance with adaptive skipping, reducing tokens by 45
FlashThink~\cite{jiang2025flashthink} uses a verifier-based early-exit strategy, cutting token usage by up to 94.7
CTS~\cite{lin2025controlling} adjusts reasoning speed in real time by editing internal representations, improving the efficiency-accuracy tradeoff.

Verifier-based and answer-aware methods further improve compression:  
VeriThinker~\cite{chen2025verithinker} trains models on auxiliary verification tasks to guide reasoning compression, significantly reducing token usage while preserving or improving accuracy.  
TrimR~\cite{lin2025trimr} uses a verifier-based pruning mechanism to detect and remove redundant reasoning steps during inference, significantly improving test-time efficiency.  
Answer Convergence~\cite{liu2025answer} applies inference-time early stopping based on convergence of predicted answers, enabling significant token reduction without compromising solution correctness.  
CTS~\cite{yuan2025not} enhances reasoning efficiency by retaining only essential tokens in chain-of-thought traces, reducing inference cost while maintaining accuracy.

Several methods employ step-level or token-level importance scoring:  
LIMOPro~\cite{xiao2025limopro} applies perplexity-based reasoning refinement to prune low-importance steps, enabling more efficient and accurate generation across complex benchmarks.  
LightThinker \cite{zhang2025lightthinker} introduces special tokens that trigger the model to dynamically compress its ongoing thought process, reducing redundancy.  
Activation-Steered Compression (ASC)~\cite{azizi2025activation} injects a learned activation vector during inference to modulate internal states, enabling concise and math-focused rationales without additional training or accuracy loss.  
TALE-EP \cite{TALE} dynamically adjusts the allotted reasoning tokens depending on task complexity.  
Meta-Reasoner \cite{meta_reasoner} applies a contextual multi-armed bandit to optimize efficiency.  
SelfBudgeter~\cite{li2025selfbudgeter} adaptively estimates token budgets based on problem complexity and enforces budget adherence during reasoning, reducing output length without sacrificing accuracy.

Memory and representation-level pruning also offer notable benefits:  
RPC~\cite{song2025reasoning} compresses reasoning paths by periodically pruning the KV cache based on inherent semantic sparsity, achieving up to 4× memory reduction and 1.6× speedup.  
Prune-on-Logic~\cite{zhao2025can} constructs logic graphs from Long Chain-of-Thought (Long-CoT) traces and selectively prunes low-utility reasoning steps under well-defined semantic constraints, enabling more efficient and accurate inference in resource-limited small language models.

Other methods optimize reasoning strategies:
Dynamic Thinking~\cite{xu2025scalable} reduces overthinking and improves efficiency by segment-level pruning and preference-based learning.
Causal~\cite{yu2025causal} prunes redundant steps in CoT reasoning using probabilistic causal processes, enhancing efficiency without losing accuracy.
DRP~\cite{jiang2025drp} achieves token efficiency gains by combining pruning with skill-aware decomposition and distillation, without accuracy loss.
ReCUT~\cite{jin2025recut} balances reasoning depth and brevity using long-short switched sampling and parameter interpolation, with minimal performance degradation.
R1-Compress~\cite{wang2025r1} reduces token usage via a two-stage chunk-level compression strategy, preserving coherence.
A*-Thought~\cite{xu2025a} compresses reasoning chains using bidirectional A* search guided by token-level importance, improving the accuracy-efficiency tradeoff.

\begin{figure}[ht]
\centering
\small
\begin{minipage}{1.0\linewidth}
\centerline{\includegraphics[width=1\textwidth]{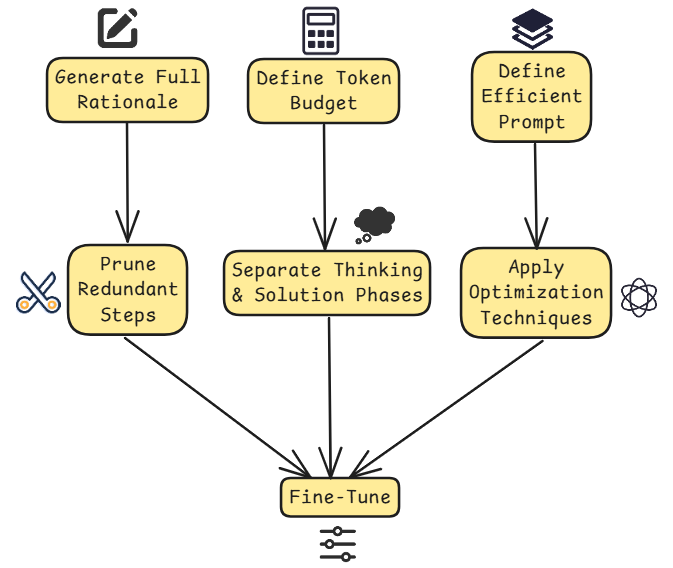}}
\end{minipage}
\caption{\textbf{Flowchart of Fine-Tuning on Compact Reasoning Chains.} Each column represents one kind of strategy of SFT for token efficiency. }
\label{fig:fine_tuning_cot}
\end{figure}

\subsubsection{Fine-Tuning on Compact Reasoning Chains}

As shown in Figure~\ref{fig:fine_tuning_cot}, fine-tuning on compact reasoning data enables LRMs to internalize efficient inference behaviors while keeping performance across diverse tasks.

Several methods generate or use condensed versions of chain-of-thought (CoT) reasoning data:  
C3oT \cite{kang2024c3ot} leverages an LLM to generate condensed versions of long CoTs, preserving essential structure before jointly training models on both full and compressed chains.  
Skip Steps \cite{liu2024can} curates expert-validated answers with condensed steps and fine-tunes LLMs to mimic these concise reasoning paths.  
SOLAR \cite{li2025solar} fine-tunes LLMs using datasets annotated for both correctness and the effectiveness of the underlying task-specific reasoning topology, encouraging minimal yet truly complete logic flows with consistent performance.

To prune redundancy in reasoning, some works focus on rationale reduction:  
VARR~\cite{jang2025varr} proposes a sentence-level rationale reduction framework guided by verbosity likelihood to prune redundant reasoning steps, significantly improving efficiency while preserving accuracy on arithmetic and commonsense tasks.  
TokenSkip \cite{xia2025tokenskip} prunes reasoning chains token-by-token based on importance, followed by fine-tuning across various compression ratios to balance brevity and precision.  
SmartThinker~\cite{he2025smartthinker} employs a two-stage framework that combines supervised fine-tuning and reinforcement learning with step-level importance-aware compression, selectively preserving essential reasoning steps while removing redundant ones.

From the perspective of controlling token usage during inference or fine-tuning:  
TALE-EP \cite{TALE} enhances token-budget awareness via SFT and direct preference optimization (DPO).  
Elastic Reasoning~\cite{xu2025scalable} separates the reasoning process into thinking and solution phases with explicit token budgets, enabling efficient CoT generation under strict inference-time constraints.  
CoT-Valve \cite{ma2025cot} discovers a latent direction that controls reasoning length, enabling models to flexibly adjust their level of detail based on task demands.

Some works avoid fine-tuning and use lightweight or prompt-based approaches:
PREMISE~\cite{yu2025premise} introduces a prompt-only framework for multi-objective optimization, balancing brevity and correctness to reduce token usage without fine-tuning.
\textsc{L2}\cite{chen2025less} combines high-quality English samples with multilingual CoTs and a lightweight decoding intervention, achieving long reasoning with reduced token cost.
EfficientXLang\cite{ahuja2025efficientxlang} shows that reasoning in non-English languages can reduce token consumption without performance loss, offering a promising multilingual strategy.
ConciseHint~\cite{tang2025concisehint} injects concise, task-adaptive hints during generation, reducing token usage while maintaining accuracy on multiple benchmarks.
Budget Guidance~\cite{li2025budget} uses a lightweight controller to adjust reasoning length during inference, achieving controlled token usage with maintained or improved accuracy.

Finally, several methods explore mixing long and short CoT data during training:  
LS-Mixture SFT~\cite{yu2025long} fine-tunes models on a mixture of long and short chain-of-thought data, promoting efficient reasoning while reducing unnecessary overthinking.  
TLDR~\cite{li2025tldr} proposes a dynamic re-weighting strategy for mixing short and long chain-of-thought data during training, enabling models to generalize across diverse reasoning lengths and achieving substantial compression (~40\%) without compromising overall performance on math reasoning benchmarks.

\subsubsection{Reward-Based Incentivization}

A growing body of work introduces explicit reward signals to effectively reduce unnecessary CoT complexity while preserving high accuracy across diverse tasks.
However, recent studies on LLM-based preference evaluation \cite{hu2024rethinking} have highlighted inherent biases in automatic preference scoring, which may also affect the reliability of CoT-length optimization objectives.

\begin{figure}[ht]
\centering
\small
\begin{minipage}{1.0\linewidth}
\centerline{\includegraphics[width=1\textwidth]{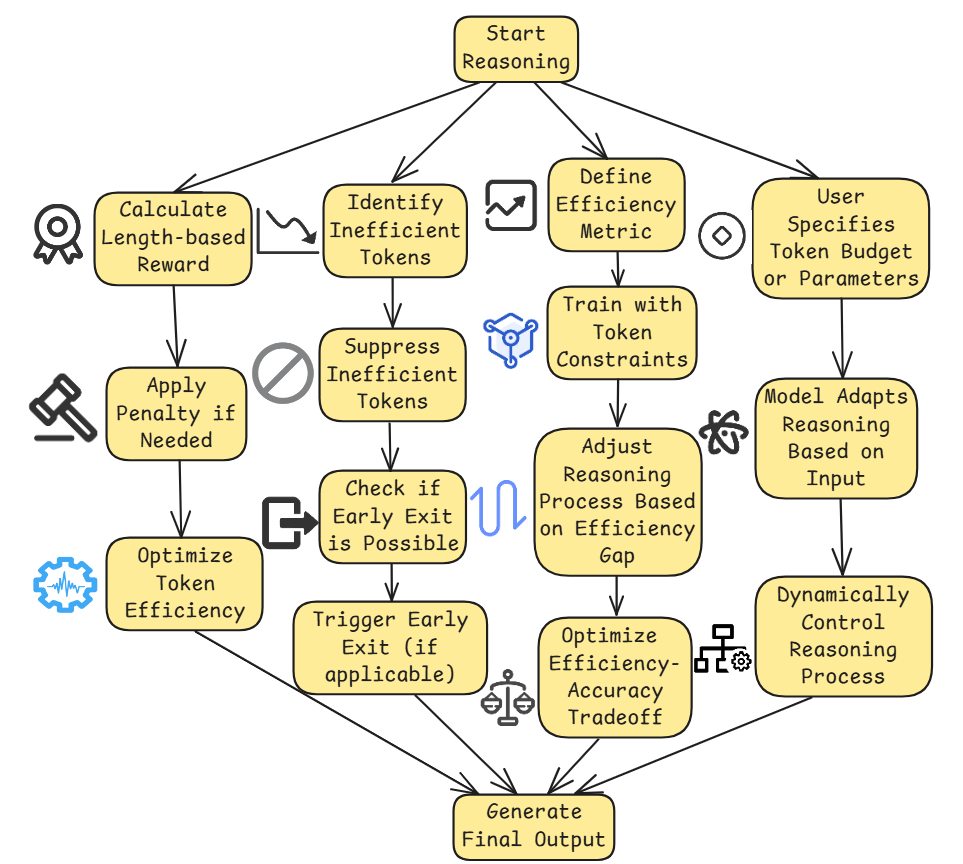}}
\end{minipage}
\caption{\textbf{Flowchart of Reward-Based Incentivization.} Each column represents one distinct kind of approach for incentivizing the token efficiency, highlighting the key steps of each method.}
\label{fig:reward_based_incentivization}
\end{figure}

The methods for Reward-Based Incentivization are illustrated in Figure~\ref{fig:reward_based_incentivization}. The flowchart highlights how different strategies, such as length-based rewards, harmonizing penalties, and reinforcement learning (RL) techniques, contribute to improving token efficiency in reasoning.

\begin{table*}[!t]
\centering
\setlength{\tabcolsep}{3pt}
\caption{\label{tab:taxonomy_3_2}\textbf{Taxonomy of Explicit Compact CoT Methods (Part II).} The criteria mainly contain training, strategy, model, and application. }
\resizebox{\linewidth}{!}{
\begin{tabular}{lccccc}
\toprule[1.1pt]
\textbf{Types} & \textbf{Methods}  & \textbf{Training} & \textbf{Strategy} & \textbf{Model} & \textbf{Application} \\
\midrule
\multirow{60}{*}{\shortstack[l]{Explicit \\ Compact \\CoT}} 
& SelfBudgeter \cite{li2025selfbudgeter} & \checkmark & RL & DeepSeek-R1-Distill-Qwen-1.5B & Math\\
& Long Short \cite{ning2025not} & \checkmark & SFT, RL & Qwen2.5-7B, Llama3.1-8B & Math, Logical\\
& Length-Aware Optimization \cite{Danlong2025EfficientRL} & \checkmark & RL & Qwen-2.5-7B & Math, Logic\\
& Prune-on-Logic \cite{zhao2025can} & \checkmark & SFT & \makecell{DeepSeek-R1-Distill-Llama-8B, \\DeepSeek-R1-Distill-Qwen7B} & Math, Logic\\
& DRP \cite{jiang2025drp} & \checkmark & SFT & DeepSeek-R1-Distill-Qwen-7B/1.5B & Math\\
& FlashThink \cite{jiang2025flashthink} & \checkmark & Prompt, SFT & \makecell{Qwen2.5, Llama-3.1-8B-Instruct, \\Mistral-7B-Instruct-v0.3, Qwen3}  & Math, Reasoning\\
& AnytimeReasoner \cite{Qi2025OptimizingAR} & \checkmark & RL & DeepSeek-R1-Distill-Qwen-1.5B & Math \\
& VeriThinker \cite{chen2025verithinker} & \checkmark & SVFT & \makecell{DeepSeek-R1-Distill-Qwen-7B/14B, \\DeepSeek-R1-Distill-Llama-8B} & Math, Reasoning\\
& LASER \cite{liu2025learn} & \checkmark & RL & DeepSeek-R1-Distill-Qwen-1.5B/7B/32B & Math, Reasoning, Code\\
& R1-Compress \cite{wang2025r1} & \checkmark & SFT & Qwen2.5-14B/32B-Instruct & Math, Logic, Scientific \\
& ACPO \cite{cheng2025incentivizing} & \checkmark & SFT, RL & \makecell{DeepSeek-R1-Distill-Qwen-1.5B/7B, \\DeepSeek-R1-Distill-Llama-8B} & Math\\
& ConciseRL \cite{Dumitru2025ConciseRL} & \checkmark & RL & \makecell{DeepSeek-R1-Distill-Qwen-1.5B, \\STILL-3-1.5B-preview} & Math, Commonsense\\
& CTS \cite{yuan2025not} & \checkmark & STF & Qwen2.5-7B/14B & Math\\
& PIR \cite{xiao2025limopro} & \checkmark & SFT & Qwen-2.5-32B & Math, Science\\
& ConciseR \cite{song2025walk} & \checkmark & RL &  Qwen2.5-Math-7B & Math\\
& CoThink \cite{fan2025cothink} & \checkmark & SFT, RL, Distillation & \makecell{Qwen2.5-Instruct-32B, \\DAPO, DeepSeek-R1-Distill, QwQ} & Math\\
& DTO \cite{an2025think} & \checkmark &  SimPO & \makecell{DeepSeek-R1-Distill-Qwen-1.5B, \\DeepScaleR-1.5B-Preview, \\Llama - 3.3 - 70B - Instruct} & Math, Reasoning\\
& A*-Thought \cite{xu2025a} & \checkmark & SFT & \makecell{QwQ-32B, \\DeepSeek-R1-Distill-Qwen-32B, \\s1.1-32B} & Math\\
& TLDR \cite{li2025tldr} & \checkmark & SFT, RL & DeepSeek-R1-Distill-7B/14B & Math \\
& Answer Convergence \cite{liu2025answer} & \checkmark & Inference-time & Qwen-32B, Qwen-7B, Llama-8B/70B, QwQ-32B & Math\\
& REO-RL \cite{gao2025how} & \checkmark & RL & DeepSeek-R1-Distill-Qwen, Qwen3 & Math \\
& Overclocking LLM Reasoning \cite{eisenstadt2025overclocking} & \checkmark & Intervention & DeepSeek-R1-LLaMA-8B/Qwen-32B & Math \\
& BINGO \cite{liu2025bingo} & \checkmark & SFT, RL & Qwen-1.5B / Qwen-7B / Qwen2.5-Math-7B & Math \\
& Brevity \cite{poddar2025brevity} & \checkmark & Prompt & GPT-3.5, Llama-2/3, Gemma, Mistral, Phi-3, Falcon, Vicuna & \makecell{Commonsense, Logic, Scientific,\\ Language Understanding, Instruction} \\
& NoWait \cite{wang2025wait} & \checkmark & Inference-Time Filtering & Qwen-3-32B, Phi4, QwQ, Kimi-VL, QvQ & \makecell{Math, Logic, Scientific, \\Commonsense, Code, Multimodal} \\
& Causal \cite{yu2025causal} & \checkmark & SFT, RL & Llama-3.2-1B-Instruct/Qwen-1.5B & Math, Commonsense \\
& PREMISE \cite{yu2025premise} & \checkmark & Prompt & Claude-3.7-Sonnet / GPT o1 / Gemini-2.5 & Math \\
& Budget Guidance \cite{li2025budget} & \checkmark & Inference-Time Guidance & DeepSeek-R1-Distill-Qwen-7B/32B, Qwen3-8B & Math, Logic, Scientific, Code \\
& ReCUT \cite{jin2025recut} & \checkmark & SFT, RL & Llama-3.1-8B-Instruct/Qwen2.5-7B-Instruct & Math \\
& PLP \cite{ling2025fast} & \checkmark & SFT, RL & DeepSeek-R1-Distill-Qwen-1.5B/7B, Qwen2.5-7B-Instruct & Math \\
& SReF \cite{liu2025efficient} & \checkmark & SFT, RL & R1-Distill-Qwen-1.5B/7B/32B, QwQ-32B, Qwen3-32B & Math \\
& LC-R1 \cite{cheng2025optimizing} & \checkmark & SFT, RL & DeepSeek-R1-Distill-Qwen-1.5B/7B & Math, Code \\
& CoLE \cite{zhao2025efficient} & \checkmark & SFT, RL & Llama-3.2-1B-Instruct/Qwen-1.5B & Math \\
& ConciseHint \cite{tang2025concisehint} & \checkmark & SFT, RL & DeepSeek-R1/Qwen3-1.7B/4B/8B & Math, Science \\
& AdapThink \cite{wan2025adapthink} & \checkmark & SFT, RL & DeepSeek-R1-Distill-Qwen-1.5B & Math \\
& \textsc{L2} \cite{chen2025less} & \checkmark & SFT, Decoding Intervention & Qwen2.5-32B & Math, Science\\
& DuP-PO \cite{ding2025thinking} & \checkmark & RL & DeepSeek-R1-Distill-Qwen-1.5B & Math \\
& AALC \cite{li2025aalc} & \checkmark & RL & Qwen2.5-Math-7B, DeepSeek-R1-Distill-Qwen-7B & Math \\
& EfficientXLang \cite{ahuja2025efficientxlang} & \ding{55} & Prompt & DEEPSEEK R1, QWEN 2.5, QWEN 3 & Math \\
& ASC \cite{azizi2025activation} & \checkmark &Inference-time & DeepSeek-R1-Distill-LLaMA-8B, Qwen-7B, QwQ-32B & Math \\
& SmartThinker \cite{he2025smartthinker} & \checkmark & SFT, RL & DeepSeek-R1-Distill-Qwen-1.5B / 7B & Math, Reasoning \\
& CTS \cite{lin2025controlling} & \checkmark & None (Plug-and-play) & \makecell{DeepSeek-R1-Distill-Qwen-7B, DeepSeek-R1-Distill-Qwen-32B,\\ QwQ-32B, Qwen3-8B} & Math, Science, Code \\
& VARR \cite{jang2025varr} & \checkmark & SFT & Mistral-7B/Llama-3.2-1B-3B & Math, Commonsense \\
\bottomrule[1.1pt]
\end{tabular}}
\end{table*}

Several works introduce length-based or harmonizing reward mechanisms:  
Kimi k1.5 \cite{team2025kimi} integrates length-based rewards to discourage verbose reasoning.  
O1-Pruner \cite{luo2025o1} detects "length disharmony" and applies harmonizing penalties that promote brevity without sacrificing solution quality.  
TLDR~\cite{zhang2025making} combines temperature scaling with length-regularized reinforcement learning to improve token efficiency in small language models without compromising reasoning accuracy on math benchmarks.  
Arora et al. \cite{arora2025training} use reinforcement learning to train models that dynamically allocate computational resources based on task difficulty, balancing cost and precision.  
DAST \cite{DAST} proposes a Token Length Budget metric that aligns task complexity with output length, encouraging efficiency through targeted penalties and rewards.  
PLP~\cite{ling2025fast} introduces a reward-modulated reinforcement learning framework that adaptively penalizes output length based on task difficulty, enabling more concise responses for simple tasks while preserving depth on challenging high-complexity reasoning tasks.

While length penalties are widely used to encourage brevity, \cite{hu2024explaining} reveals that LLM-based preference evaluations can exhibit a systematic length bias, favoring unnecessarily long responses in pairwise comparisons.
This bias implies that naive length penalties or rewards must be carefully designed to avoid counteracting model alignment goals.

Other methods refine reward structures using token-level semantics or inefficiency suppression: 
DuP-PO~\cite{ding2025thinking} introduces Dual-Policy Preference Optimization, a reinforcement learning strategy that suppresses inefficient "thinking tokens" (e.g., wait, however), improving both accuracy and token efficiency in math-focused LLMs.  
S-GRPO~\cite{dai2025sgrpo} applies a decaying-reward reinforcement learning strategy to encourage early exits in reasoning chains, reducing token usage by up to 61\% while improving accuracy across math and science tasks.  
BINGO~\cite{liu2025bingo} introduces dynamic significance-aware reward signals for CoT length optimization under an RL framework, enhancing token efficiency without compromising performance.  
IBPO \cite{yu2025think} adopts a constrained RL framework to control the distribution of reasoning across response groups based on inference cost.  
AdapThink~\cite{wan2025adapthink} applies confidence-aware and diversity-sensitive reinforcement learning to dynamically regulate reflection and reasoning depth, improving both efficiency and accuracy in complex reasoning tasks.

\begin{table*}[!t]
\centering
\setlength{\tabcolsep}{3pt}
\caption{\label{tab:taxonomy_4}\textbf{A Taxonomy of Efficient Inference Methods for Large Reasoning Models.} The criteria mainly contain training, strategy, model, and application.}
\resizebox{\linewidth}{!}{
\begin{tabular}{lccccc}
\toprule[1.1pt]
\textbf{Types} & \textbf{Methods}  & \textbf{Training} & \textbf{Strategy} & \textbf{Model} & \textbf{Application} \\
\midrule
\multirow{12}{*}{\shortstack[l]{Implicit \\ Latent \\CoT}} 
& Soft Thinking \cite{zhang2025softthinking} & \ding{55} & Decoding & Qwen-32B/70B, LLaMA-70B & Math, Code \\
&  
ICoT-KD~\cite{deng2023implicit}  & \checkmark & SFT & GPT-2 Small/Medium & Math\\ 
& CODI~\cite{CODI}  & \checkmark & SFT & GPT-2 Small, LLaMA-3.2-1B & Math\\
& ICoT-SI~\cite{deng2024explicit}  & \checkmark & SFT & GPT-2 Small/Medium, Phi-3 3.8B, Mistral 7B & Math\\
& COCONUT~\cite{hao2024training}  & \checkmark & SFT & GPT-2 & Math\\
& CCoT~\cite{cheng2024compressed}  & \checkmark & SFT & LLaMA2-7B-Chat & Math, Logic\\
& Heima \cite{shen2025efficient} & \checkmark & SFT &
LLaVA-CoT, LLaMA-3.1-8B-Instruct &  Multimodal Reasoning\\
& Token assorted \cite{su2025token} & \checkmark & SFT &
LLaMA-3.2-1B, LLaMA-3.2-3B, LLaMA-3.1-8B &
Agentic Planning, Logic, Math. \\
& SoftCoT \cite{xu2025softcot}  & \checkmark & SFT &
LLaMA-3.1-8B-Instruct, Qwen2.5-7B-Instruct &
Math, Commonsense,Reasoning \\

& CoLaR \cite{tan2025think} & \checkmark & SFT, RL & Llama-3.2-1B-Instruct/Qwen-1.5B & Math \\
& Efficient Latent Refinement \cite{wang2025efficient} & \checkmark & Post-training (training-free) & LLaMA-3.2-3B / Qwen-2.5-1.5B / GPT-2 & Math, Commonsense, Multi-hop \\
& DART \cite{jiang2025dart} & \checkmark & SFT & Llama-3.2-1B-Instruct/Qwen2.5-1.5B/GPT2 & Math\\

\bottomrule[1.1pt]
\end{tabular}}
\end{table*}

Some works propose novel training metrics or frameworks:  
REO-RL~\cite{gao2025how} defines a Reasoning Efficiency Gap (REG) metric and trains models via reinforcement learning to close this gap under token constraints, achieving improved efficiency-accuracy tradeoffs.  
CoLE~\cite{zhao2025efficient} integrates Efficiency Steering and Self-Rewarded Efficiency RL to guide large reasoning models toward shorter solution paths by leveraging their intrinsic reasoning structure.  
MRT \cite{qu2025optimizing} applies meta-reinforcement learning to balance exploration of novel reasoning paths with the exploitation of concise, proven ones.  
Short-RL~\cite{Danlong2025EfficientRL} applies length-aware reinforcement learning to reduce reasoning length by up to 40\% without extra training stages, maintaining strong performance on logic and math tasks.  
LASER and its adaptive variants LASER-D/DE~\cite{liu2025learn} use reinforcement learning with difficulty-aware reward shaping to balance reasoning accuracy and token efficiency through adaptive length control.

Interactive and user-directed length control mechanisms are also emerging:  
Claude 3.7 \cite{claud3_7}, the first hybrid reasoning model, introduces an extended thinking mode where users can prescribe token budgets.  
ACPO~\cite{cheng2025incentivizing} integrates dual-process reasoning and difficulty-aware length budgeting into an RL framework, enabling dynamic cognitive control and efficient token use in complex tasks.  
L1 \cite{l1} generalizes this idea with Length Controlled Policy Optimization (LCPO), enabling fully configurable CoT lengths at inference time.  
AnytimeReasoner~\cite{Qi2025OptimizingAR} uses budget-relative policy optimization to guide reasoning under variable token limits, enabling adaptive token usage without accuracy degradation.  
Overclocking LLM Reasoning~\cite{eisenstadt2025overclocking} leverages learned internal progress vectors to monitor and accelerate reasoning phases in real time, improving efficiency and interpretability.  
Long Short~\cite{ning2025not} uses a collaborative multi-turn reinforcement learning setup, where specialized LLMs for long and short thoughts jointly compress reasoning chains, reducing token usage while maintaining high accuracy.

Other innovative strategies further improve reward-guided compression:  
ConciseRL~\cite{Dumitru2025ConciseRL} leverages an LLM-judged conciseness reward in a hyperparameter-free RL setting to train models for succinct and accurate reasoning.  
Brevity~\cite{poddar2025brevity} analyzes verbosity in LLM responses and proposes prompt engineering techniques to reduce reasoning length, enhancing energy efficiency without sacrificing accuracy.  
ConciseR~\cite{song2025walk} adopts a two-stage reinforcement learning approach that first ensures correctness, then compresses reasoning to optimize length without sacrificing performance.  
LC-R1~\cite{cheng2025optimizing} combines length- and compression-based rewards within a GRPO framework to eliminate invalid reasoning patterns, achieving approximately 50\% output compression with minimal accuracy loss across diverse reasoning benchmarks.  
AALC~\cite{li2025aalc} proposes an accuracy-aware length reward to guide LLMs toward balancing brevity and correctness, reducing response length by over 50\% while maintaining high reasoning accuracy.

\subsubsection{Takeaways of Explicit Compact CoT}

We distill several important insights from our analysis of Explicit Compact CoT strategies. These takeaways reflect critical aspects of reasoning transparency, dataset constraints, reward optimization, and practical deployment challenges.

\begin{tcolorbox}[colframe=black, colback=gray!5, coltitle=white, title=Takeaways of Explicit Compact CoT]
\begin{itemize}[left=0pt, label=\textbullet, itemsep=1.2em]

    \item CoT compression enhances scalability but may sacrifice transparency. These techniques lower token usage by abstracting reasoning steps, but risk omitting essential intermediate logic, which can undermine interpretability.

    \item Supervised fine-tuning improves efficiency, but at high cost. While effective, these methods depend on curated, condensed datasets and heavy preprocessing, limiting their adaptability to open-ended domains.

    \item Reward-based brevity can lead to shallow reasoning. Incentivizing shorter outputs may cause models to favor simplistic answers, at the expense of the deeper reasoning needed for complex tasks.

    \item Efficiency alone is insufficient for real-world deployment. Real-world applications require a balance between compactness and reasoning robustness, interpretability, and domain generalization.

\end{itemize}
\end{tcolorbox}

\subsection{Implicit Latent CoT}
\label{sec:implicit-cot}

Implicit latent CoT methods boost token efficiency by shifting reasoning \textbf{from explicit tokens to latent tokens}, encoding reasoning in hidden layers rather than natural language.


A line of knowledge distillation methods~\cite{deng2023implicit,deng2024explicit,CODI} trains student models to infer the teacher's internal CoT representations rather than mimic explicit token sequences, enabling ``vertical'' reasoning across transformer layers. Chain of Continuous Thought (COCONUT)\cite{hao2024training} replaces token-level reasoning chains with autoregressively generated latent embeddings, which are then fed back into the model to emulate breadth-first search during complext problem-solving. Compressed CoT (CCoT)\cite{cheng2024compressed} introduces contemplation tokens—dense, compressed representations of full reasoning chains—significantly reducing inference latency while maintaining high accuracy.

\begin{table}[!t]
\centering
\setlength{\tabcolsep}{6pt}
\caption{\label{tab:benchmark_explicit_1}\textbf{Benchmarks Used by Explicit Compact CoT Methods.}}
\resizebox{\linewidth}{!}{
\begin{tabular}{lcc}
\toprule[1.1pt]
\textbf{Types} & \textbf{Methods} & \textbf{Application (Benchmarks)} \\
\midrule
\multirow{80}{*}{\shortstack[l]{Explicit \\ Compact \\CoT}} 
& SoT \cite{aytes2025sketch} & MATH, CommonsenseQA, StrategyQA, ECQA \\
& Constrained-CoT \cite{nayab2024concise} & GSM8K, AQuA, SVAMP, MathQA \\
& CoD \cite{xu2025chain} & GSM8K, SVAMP, MultiArith, GSM-HARD \\
& TALE-EP \cite{TALE} & GSM8K, MATH \\
& Meta-Reasoner \cite{meta_reasoner} & Game of 24, TheoremQA, SciBench \\
& TS \cite{zhang2025making} & MATH500, AMC, AIME24, OlympiadBench \\
& Fractured Sampling \cite{liao2025fractured} & MATH500 L5, AIME24, AIME25, AIMO2, GPQA Diamond \\
& RPC \cite{song2025reasoning} & DROP, GSM8K, PRM800k, PRM12K \\
& ThinkLess \cite{li2025thinkless} & GSM8K, MMLU, GPQA, BBH \\
& PLAN-AND-BUDGET \cite{lin2025plan} & GSM8K, DROP, ARC \\
& TrimR \cite{lin2025trimr} & MATH500, AIME24, AIME25, GPQA Diamond \\
& SOLAR \cite{li2025solar} & GSM8K, MATH \\
& C3oT \cite{kang2024c3ot} & GSM8K, MathQA, ECQA, StrategyQA \\
& TokenSkip \cite{xia2025tokenskip} & GSM8K, MATH500 \\
& InftyThink \cite{yan2025inftythink} & MATH500, AIME24, GPQA Diamond \\
& LightThinker \cite{zhang2025lightthinker} & GSM8K, MMLU, GPQA, BBH \\
& CoT-Valve \cite{ma2025cot} & GSM8K, AIME24, PRM800k, PRM12K \\
& Distill System 2 \cite{yu2024distilling} & \makecell{Last Letter Concatenation, Coin Flip, SycophancyEval, \\ OASST2, MT-Bench, GSM8k} \\
& SF \cite{munkhbat2025self} & GSM8K, MATH \\
& Skip Steps \cite{liu2024can} & Analog of Algebra, Multi-digit Addition, Directional Reasoning \\

& DAST \cite{DAST} & AIME24, AIME25, AMC2023, MinervaMATH, MATH500 \\
& TALE-PT \cite{TALE} & GSM8K, MATH \\
& Kimi k1.5 \cite{team2025kimi} & \makecell{MMStar, MMBench V1.1, MMVet, \\ MathVista, AI2D, HallusionBench} \\
& O1-Pruner \cite{luo2025o1} & AIME, AMC, GPQA Diamond \\
& MRT \cite{qu2025optimizing} & AIME2024, AIME2025, AMC2023, MinervaMATH, MATH500 \\
& ERL \cite{arora2025training} & GSM8K, MATH500, AIME2024, CommonsenseQA, Logical Deduction \\
& Claude 3.7 \cite{claud3_7} & GSM8K, BIG-bench, Coin Flip, MathBench \\
& L1 \cite{l1} & AIME2025, AMC, MATH, OlympiadBench, GPQA, LSAT, MMLU \\
& SPIRIT \cite{cui2025stepwise} & \makecell{Algebra-Linear-1d, Number-Base-Conversion, \\ Diff-Calc, Time-Diff, GSM8K, MetaMathQA} \\
& IBPO \cite{yu2025think} & MATH500, AMC, Qsdpo, Asdpo\_golden \\
& LS-Mixture SFT \cite{yu2025long} & MATH500, AIME24, GPQA Diamond \\
& ConCISE \cite{qiao2025concise} & GSM8K, Math-500, AIME24, GPQA Diamond \\
& Elastic Reasoning \cite{xu2025scalable} & AIME2024, AMC, MATH500, OlympiadBench, Minerva Math \\
& S-GRPO \cite{dai2025sgrpo} & GSM8K, AIME2024, AMC2023, MATH-500, GPQA Diamond \\
& TLDR \cite{zhang2025making} & MATH500, AMC, AIME24, OlympiadBench \\
& Adaptive GoGI-Skip \cite{zhuang2025accelerating} & AIME2025, AIME2024, GPQA Diamond, GSM8K \\
& SelfBudgeter \cite{li2025selfbudgeter} & GSM8K, MATH, AIME2024 \\
& Long Short \cite{ning2025not} & MATH500, AIME2024, AIME2025, AMC2023, GPQA Diamond \\
& Length-Aware Optimization \cite{Danlong2025EfficientRL} & \makecell{Logic-RL dataset, AMC23, AIME2024, \\ MATH500, Minerva Math, Olympiad Bench} \\
& Prune-on-Logic \cite{zhao2025can} & AMC23, AIME, MATH500, GSM8K, BBH \\
& DRP \cite{jiang2025drp} & GSM8K, PRM12K, MATH500, AIME24, AMC23 \\
& FlashThink \cite{jiang2025flashthink} & GSM8K, MATH, GPQA Diamond, DROP \\
& AnytimeReasoner \cite{Qi2025OptimizingAR} & AIME2024, AMC2022, MATH500, Minerva Math, OlympiadBench \\
& VeriThinker \cite{chen2025verithinker} & MATH500, AIME2024, AIME2025, GSM8K \\
& LASER \cite{liu2025learn} & MATH500, AIME2024, AMC2023, OlympiadBench, GPQA, MMLU, LSAT \\
& R1-Compress \cite{wang2025r1} & MATH500, AIME24, GPQA Diamond \\
& ACPO \cite{cheng2025incentivizing} & MATH500, AIME2024, GSM8K \\
& ConciseRL \cite{Dumitru2025ConciseRL} & GSM8K, MATH500, TheoremQA, GPQA-main, MMLU-Pro-1k \\
& CTS \cite{yuan2025not} & MATH500, AIME24, GPQA Diamond \\
& PIR \cite{xiao2025limopro} & AIME, AMC, GPQA Diamond \\
& ConciseR \cite{song2025walk} & AIME2024, MATH-500, AMC2023, Minerva, OlympiadBench \\
& CoThink \cite{fan2025cothink} & GSM8K, MATH500, AIME24 \\
& DTO \cite{an2025think} & GSM8K, MATH500, Gaokao, AMC2023, AIME2024, AIME2025 \\
& A*-Thought \cite{xu2025a} & MATH500, AMC23, OlympiadBench, GSM8K \\
& TLDR \cite{li2025tldr} & GSM8K, MATH, AIME, AMC, ASDiv, Minerva \\
& Answer Convergence \cite{liu2025answer} & NQ, GSM8K, MATH-500, GPQA, AIME'24 \\
& REO-RL \cite{gao2025how} & AMC 2023, AIME 2024, AIME 2025, Minerva Math \\
& Overclocking LLM Reasoning \cite{eisenstadt2025overclocking} & GSM8K, Math500 \\
& BINGO \cite{liu2025bingo} & GSM8K, MATH500, TheoremQA, AIME2024 \\
& Brevity \cite{poddar2025brevity} & DOLLY, GOOAQ, MS-MARCO, NARRATIVEQA, TWEETQA \\
& NoWait \cite{wang2025wait} & \makecell{AMC 2023, AIME 2024, AIME 2025, GPQA-D, MMMU,\\ MMMU-Pro, MathVista, EMMA-mini, MMVU, VSI-Bench} \\
& Causal \cite{yu2025causal} & GSM8K, MATH-500, AIME, CommonsenseQA \\
& PREMISE \cite{yu2025premise} & GSM8K, SVAMP, MATH-500 \\
& Budget Guidance \cite{li2025budget} & \makecell{MATH-500, AIME-2024, AMC, \\ OlympiadBench, GPQA, FOLIO, TableBench, LiveCodeBench} \\
& ReCUT \cite{jin2025recut} & GSM8K, AMC23, AIME24, AIME25, MATH500 \\
& PLP \cite{ling2025fast} & GSM8K, MATH500, AIME2024 \\
& SReF \cite{liu2025efficient} & MATH500, AIME24, AMC23, GSM8K \\
& LC-R1 \cite{cheng2025optimizing} & AIME25, MATH500, GSM8K, AMC, Olympiad, GPQA-D, LCB \\
& CoLE \cite{zhao2025efficient} & GSM8K-Aug, GSM-Hard, SVAMP, MultiArith, MATH \\
& ConciseHint \cite{tang2025concisehint} & GSM8K, AIME24, GPQA-Diamond \\
& AdapThink \cite{wan2025adapthink} & AIME2025, AIME2024, MATH500, AMC \\
& \textsc{L2} \cite{chen2025less} & AIME24, AIME25, GPQA-Diamond, MATH500, Graduate Entrance Exam \\
& DuP-PO \cite{ding2025thinking} & MATH500, OlympiadBench, Minerva, AIME24, AIME25, AMC \\
& AALC \cite{li2025aalc} & GSM8K, MATH, AIME24, AMC24, CNMO24, GPQA \\
& EfficientXLang \cite{ahuja2025efficientxlang} & AMC23, MATH500, AIME2024, AIME2025 \\
& ASC \cite{azizi2025activation} & GSM8K, MATH500 \\
& SmartThinker \cite{he2025smartthinker} & \makecell{AIME24, AIME25, AMC23, MinervaMATH, \\ MATH, Olympiad-Bench, TruthfulQA, RACE, Live-Code-Bench} \\
& CTS \cite{lin2025controlling} & MATH-500, AIME24, AIME25, GPQA Diamond, LiveCodeBench \\
& VARR \cite{jang2025varr} & GSM8K, MathQA, TriviaQA, CommonsenseQA, StrategyQA \\
\bottomrule[1.1pt]
\end{tabular}}
\end{table}


Heima\cite{shen2025efficient} condenses CoT stages into latent thinking tokens and incorporates an explanatory prompt at the decoder stage to interpret the compressed reasoning. SoftCoT \cite{xu2025softcot} utilizes a small instruction-tuned 1B model to obtain instance-specific latent thought tokens and trains a projection layer to incorporate thought tokens into LLM input. Soft Thinking~\cite{zhang2025softthinking} replaces discrete reasoning tokens with probabilistically weighted concept tokens, enabling reasoning in a continuous concept space without training, and improving both accuracy and token efficiency on math and code tasks. Token-Assorted CoT\cite{su2025token} mixes latent and text tokens, encoding the initial part of the CoT into VAE-based discrete latent tokens while preserving the remainder as natural language, resulting in a hybrid representation that enhances reasoning efficiency. CoLaR~\cite{tan2025think} dynamically compresses reasoning into latent representations using probabilistic latent prediction and reinforcement learning, enabling variable-speed inference with strong accuracy on mathematical benchmarks. Efficient Latent Refinement~\cite{wang2025efficient} proposes a training-free, lightweight post-training method that updates residual embeddings using contrastive feedback, boosting latent-space reasoning accuracy by up to 5\% on benchmarks like MathQA without modifying model weights or generating intermediate tokens. DART~\cite{jiang2025dart} enables efficient non-autoregressive reasoning by distilling CoT into evolving latent “silent thought” representations via a dual-pathway self-distillation framework.

While their implementations vary, these approaches share a common goal: optimizing inference by internalizing the reasoning process. Empirical results suggest that implicit latent CoT models can match or even surpass explicit CoT methods in reasoning accuracy while significantly reducing generation costs, proving their scalability and efficiency.

\begin{tcolorbox}[colframe=black, colback=gray!5, coltitle=white, title=Takeaways of Implicit Latent CoT]
\begin{itemize}[left=0pt, label=\textbullet, itemsep=1.2em]

    \item Implicit latent CoT improves efficiency by internalizing reasoning steps but sacrifices interpretability, making verification difficult.

    \item Different methods (e.g., knowledge distillation, latent embeddings, contemplation tokens) optimize reasoning at various levels, reducing latency while maintaining accuracy.

    \item Future work should focus on extracting human-interpretable reasoning traces from latent representations to balance efficiency and transparency.

\end{itemize}
\end{tcolorbox}



\begin{table}[!t]
\centering
\setlength{\tabcolsep}{6pt}
\caption{\label{tab:benchmark_implicit}\textbf{Benchmarks Used by Implicit Latent CoT Methods.}}
\resizebox{\linewidth}{!}{
\begin{tabular}{lcc}
\toprule[1.1pt]
\textbf{Types} & \textbf{Methods} & \textbf{Application (Benchmarks)} \\
\midrule
\multirow{18}{*}{\shortstack[l]{Implicit \\ Latent \\CoT}} 
& Soft Thinking \cite{zhang2025softthinking} & \makecell{MATH500, AIME2024, GSM8K, GPQA Diamond, \\ HumanEval, MBPP, LiveCodeBench} \\
& ICoT-KD~\cite{deng2023implicit} & BIG-Bench Arithmetic, GSM8K \\
& CODI~\cite{CODI} & GSM8k, SVAMP, GSM-HARD, MultiArith \\
& ICoT-SI~\cite{deng2024explicit} & Multi-digit Multiplication, GSM8K \\
& COCONUT~\cite{hao2024training} & GSM8k, ProntoQA, ProsQA \\
& CCoT~\cite{cheng2024compressed} & GSM8K \\
& Heima \cite{shen2025efficient} & \makecell{MMStar, MMBench V1.1, MMVet, \\ MathVista, AI2D, HallusionBench} \\
& Token assorted \cite{su2025token} & \makecell{Keys-Finding Maze, ProntoQA, ProsQA, \\MATH, GSM8K, Fresh-Gaokao-Math-2023, DeepMind-Math, \\College-Math, OlympiaBench-Math, TheoremQA} \\
& SoftCoT \cite{xu2025softcot} & \makecell{CommonsenseQA, OpenBookQA, \\GSM8K, Last Letter Concatenation} \\
& CoLaR \cite{tan2025think} & GSM8K-Aug, GSM-Hard, SVAMP, MultiArith, MATH \\
& DART \cite{jiang2025dart} & GSM8K-Aug, GSM-Hard, SVAMP, MultiArith \\
& Efficient Latent Refinement \cite{wang2025efficient} & GSM8K, MathQA, AQUA-RAT, StrategyQA, ProsQA \\

\bottomrule[1.1pt]
\end{tabular}}
\end{table}

\section{Empirical Analyses}
\subsection{Analyses on Reasoning Scenarios}

\begin{table}[!t]
\centering
\setlength{\tabcolsep}{3pt}
\caption{\textbf{Acc. and Token Costs of Explicit Compact CoT Methods.}}
\label{tab:benchmark_1}
\resizebox{1.0\linewidth}{!}{
\begin{tabular}{lccccc}
\toprule[1.1pt]
\textbf{Types} & \textbf{Methods}  & \textbf{Setting} & \textbf{Accuracy} & \textbf{Models} & \textbf{Token Costs} \\
\midrule
\multirow{80}{*}{\shortstack[l]{Explicit \\ Compact \\CoT}} 
&\multirow{4}{*}{CoD \cite{xu2025chain}} & zero-shot & 84.40\% & GPT-4o & 76.40 \\  
& & zero-shot & 65.50\% & Claude 3.5 Sonnet & 73.70 \\  
 & & few-shot & 91.10\% & GPT-4o & 43.90 \\  
 & & few-shot & 91.40\% & Claude 3.5 Sonnet & 39.80 \\  
& \multirow{3}{*}{TALE \cite{TALE}} & zero-shot, prompt & 84.46\% & GPT-4o-mini & 77.26 \\ 
&  & zero-shot, SFT & 74.11\% & LLaMA-3.1-8B-Instruct & 149.93 \\ 
&  & zero-shot, DPO & 78.41\% & LLaMA-3.1-8B-Instruct & 113.41 \\ 
& \multirow{2}{*}{C3oT \cite{kang2024c3ot}} & zero-shot & 36.92\% & LLaMA-2-Chat-7B & -\\
& & zero-shot & 47.10\% & LLaMA-2-Chat-13B & -\\
&\multirow{6}{*}{TokenSkip \cite{xia2025tokenskip}} & zero-shot, ratio=0.5  & 86.70\% & LLaMA-3.1-8B-Instruct & 113.05 \\
&  & zero-shot, ratio=0.6 & 86.10\% & LLaMA-3.1-8B-Instruct & 198.01 \\
&  & zero-shot, ratio=0.7 & 84.30\% & LLaMA-3.1-8B-Instruct & 169.89 \\
&  & zero-shot, ratio=0.8 & 82.50\% & LLaMA-3.1-8B-Instruct & 150.12 \\
&  & zero-shot, ratio=0.9 & 81.10\% & LLaMA-3.1-8B-Instruct & 129.38 \\
&  & zero-shot, ratio=1.0 & 78.20\% & LLaMA-3.1-8B-Instruct & 113.05 \\
& \multirow{4}{*}{LightThinker \cite{zhang2025lightthinker}} & zero-shot,tho. & 90.14\%  & \makecell{DeepSeek-R1-Distill-Qwen-7B} & - \\
&  & zero-shot,token & 87.11\% & \makecell{DeepSeek-R1-Distill-Qwen-7B} & - \\
&  & zero-shot,tho. & 88.25\% & \makecell{DeepSeek-R1-Distill-LLaMA-8B} & - \\
&  & zero-shot,tho. & 85.52\% & \makecell{DeepSeek-R1-Distill-LLaMA-8B} & - \\
& SF \cite{munkhbat2025self} & zero-shot  & 76.72\% & \makecell{DeepSeekMath-7B}  & 184.13 \\
& O1-Pruner \cite{luo2025o1} & few-shot & 96.50\%  & QwQ-32B & 343.00\\
& \multirow{4}{*}{FlashThink \cite{jiang2025flashthink}} & zero-shot & 93.99\% & DeepSeek-R1 & 90.91\% \\
&  & zero-shot & 92.65\% & QwQ-32B & 89.60\% \\
&  & zero-shot & 87.26\% & R1-Distill-Llama-70B & 75.73\% \\
&  & zero-shot & 88.32\% & R1-Distill-Qwen-32B & 76.35\% \\
& \multirow{2}{*}{VeriThinker \cite{chen2025verithinker}} & zero-shot & 96.1\% & \makecell{R1-Distill-Qwen-7B\\Qwen-2.5-Math-7B} & 407 \\  
&  & zero-shot & 96.6\% & \makecell{R1-Distill-Qwen-14B\\ Qwen-2.5-Math-7B} & 387 \\
& \multirow{3}{*}{FlashThink \cite{jiang2025flashthink}} 
& zero-shot & 89.0\% & QwQ-32B & 342.6 \\  
& & zero-shot & 96.6\% & QwQ-32B & 745.2 \\  
& & zero-shot & 96.6\% & QwQ-32B & 418.9 \\
& \multirow{2}{*}{ReCUT \cite{jin2025recut}} & zero-shot & 86.00\% & Qwen2.5-7B-Instruct & 704 \\  
& & zero-shot & 73.90\% & LLaMA-3.1-8B-Instruct & 823 \\
& DRP \cite{jiang2025flashthink} & zero-shot, SFT & 94.10\% & DeepSeek-R1-Distill-Qwen-7B & 328.00 \\
& A*-Thought \cite{xu2025a} & few-shot & 91.20\% & QwQ-32B & 843.69 \\
& \multirow{2}{*}{FlashThink \cite{jiang2025flashthink}} & zero-shot & 72.5\% & DeepSeek-R1-Distill-Qwen-1.5B & 16.4 \\
& & zero-shot & 80.9\% & DeepSeek-R1-Distill-Qwen-1.5B & 35.8 \\
& \multirow{2}{*}{SelfBudgeter \cite{li2025selfbudgeter}} & zero-shot, GSM-init & 76.27\% & DeepSeek-R1-Distill-Qwen-1.5B & 523.77 \\
& & zero-shot, s1k-init & 81.50\% & DeepSeek-R1-Distill-Qwen-1.5B & 662.08 \\
& \multirow{5}{*}{Constrained-CoT \cite{nayab2024concise}} & zero-shot, CCoT-15 & 31.5\% & Falcon-40b & 12.1 \\
& & zero-shot, CCoT-30 & 27.1\% & Falcon-40b & 13.2 \\
& & zero-shot, CCoT-45 & 27.6\% & Falcon-40b & 14.5 \\
& & zero-shot, CCoT-60 & 28.2\% & Falcon-40b & 14.9 \\
& & zero-shot, CCoT-100 & 27.4\% & Falcon-40b & 15.4 \\
& \multirow{3}{*}{ThinkLess \cite{li2025thinkless}} & zero-shot & 88.40\% & Qwen2.5-7B & 235.41 \\
&  & zero-shot & 92.49\% & Qwen2.5-14B & 235.32 \\
&  & zero-shot & 78.92\% & LLaMA3.1-8B & 260.74 \\
& \multirow{3}{*}{SOLAR \cite{li2025solar}} & zero-shot, topo-tuning & 84.00\% & Qwen2-VL-7B-Instruct & - \\
&  & zero-shot, topo-rewarding & 88.00\% & Qwen2-VL-7B-Instruct & - \\
&  & zero-shot, hybrid-scaling & 89.02\% & Qwen2-VL-7B-Instruct & - \\
& SF \cite{munkhbat2025self} & zero-shot, FS-Self & 77.27\% & \makecell{LLaMA-3.2-3B\\Gemma-2-2B\\Qwen2.5-3B\\Qwen2.5-Math-1.5B\\DeepSeekMath-7B} & 190.03 \\
& \multirow{4}{*}{S-GRPO \cite{dai2025sgrpo}} & zero-shot & 93.8\% & DeepSeek-R1-Distill-Qwen-7B & 906 \\
&  & zero-shot & 96.2\% & DeepSeek-R1-Distill-Qwen-14B & 724 \\
&  & zero-shot & 96.1\% & Qwen3-8B & 1,292 \\
&  & zero-shot & 96.3\% & Qwen3-14B & 952 \\
& \multirow{2}{*}{ConciseRL \cite{Dumitru2025ConciseRL}} & zero-shot & 80.9\% & DeepSeek-R1-Distill-Qwen-1.5B & 543 \\  
&  & zero-shot (Separated) & 72.5\% & DeepSeek-R1-Distill-Qwen-1.5B & 248 \\
& DTO \cite{an2025think} & zero-shot & 83.91\% & DeepSeek-R1-Distill-Qwen-1.5B & 844.18 \\
& TLDR \cite{li2025tldr} & zero-shot & 87.70\% & DeepSeek-R1-Distill-Qwen-7B & 253 \\
& \multirow{2}{*}{Overclocking \cite{eisenstadt2025overclocking}} & zero-shot, $\alpha$=100 & 85.96\% & DeepSeek-R1-Distill-Qwen-32B & $\sim$240 \\  
& & zero-shot, $\alpha$=100 & 39.87\% & DeepSeek-R1-Distill-LLaMA-8B & $\sim$340 \\
& \multirow{3}{*}{BINGO \cite{liu2025bingo}} 
& zero-shot & 87.32\% & GPT-4o & 71.40 \\  
&  & few-shot & 92.15\% & GPT-4o & 41.80 \\  
&  & zero-shot & 79.44\% & LLaMA-3.1-8B-Instruct & 120.50 \\  
& Causal \cite{yu2025causal} & zero-shot, PNS-optimized & 99.9\% & DeepSeek-V3 & 52.2 \\
& \multirow{3}{*}{PREMISE \cite{yu2025premise}} & zero-shot & 95.00\% & Claude 3.7 Sonnet & - \\
& & zero-shot & 97.00\% & OpenAI o1 & - \\
& & zero-shot & 95.00\% & Gemini 2.5 Flash & - \\
& PLP \cite{ling2025fast} & zero-shot & 90.10\% & DeepSeek-R1-Distill-Qwen-7B & 218 \\
& \multirow{3}{*}{ConciseHint \cite{tang2025concisehint}} & zero-shot, AdaP+ConciseHint & 94.75\% & Qwen3-4B & 839 \\
& & zero-shot, AdaP+ConciseHint & 95.51\% & Qwen3-8B & 935 \\
& & zero-shot, AdaP+ConciseHint & 93.31\% & DeepSeek-R1-14B & 573 \\
& \multirow{2}{*}{AALC \cite{li2025aalc}} 
& zero-shot & 97.59\% & Qwen2.5-Math-7B & 97.01 \\
& & zero-shot & 97.72\% & DeepSeek-R1-Distill-Qwen-7B & 100.58 \\
& \multirow{2}{*}{ASC \cite{azizi2025activation}} & zero-shot & 88.60\% & DeepSeek-R1-Distill-Qwen-7B & 536 \\
& & zero-shot & 89.30\% & DeepSeek-R1-Distill-LLaMA-8B & 850 \\
& \multirow{1}{*}{VARR \cite{jang2025varr}} & zero-shot & 54.98\% & Mistral 7B & 100.38 \\
\bottomrule[1.1pt]
\end{tabular}}
\end{table}

This section conducts empirical analyses of existing reasoning-efficient methods from the perspectives of both performance and token efficiency. In this subsection, we examine the benchmarks adopted in prior work, focusing on their coverage across diverse reasoning scenarios and their implications for performance evaluation. To provide a structured view, the surveyed benchmarks are categorized into ten representative reasoning scenarios, each reflecting distinct task characteristics and cognitive demands.
\begin{enumerate}[label=\textbullet, leftmargin=0.4cm, itemsep=0.2em, parsep=0.2em, topsep=0.em]  
    \item \textbf{Mathematical Reasoning}: This category encompasses datasets from grade-school arithmetic (GSM8K \cite{cobbe2021gsm8k}, GSM8K-Zero \cite{chiang2024reasoning}, SVAMP \cite{patel2021nlp}, AQuA \cite{ling2017program}, ASDiv \cite{miao2020diverse}) to advanced high-school and competition-level mathematics (MathBench \cite{liu2024mathbench}, TheoremQA \cite{chen2023theoremqa}, MATH \cite{hendrycksmath2021}, MathQA \cite{yu2023metamath}, AIME24 \cite{AIME}, Olympiad-Bench \cite{he2024olympiadbench}) and graduate-level STEM reasoning (GPQA \cite{rein2024gpqa}). Collectively, they evaluate multi-step quantitative reasoning, linguistic robustness, and problem-solving depth across diverse mathematical domains.

    \item \textbf{Causal Reasoning}: Encompasses datasets such as QASC \cite{Khot2019QASC} and WorldTree \cite{jansen2018worldtree}, which test the ability to identify and link underlying cause–effect relationships, often through multi-hop scientific reasoning.
    
    \item \textbf{Code Reasoning}: Includes LiveCodeBench~\cite{jain2024livecodebench}, Codeforces, and SWE-bench~\cite{jimenez2023swe}, evaluating program synthesis, code understanding, and bug fixing in coding environments with strong practical relevance.
    
    \item \textbf{Logical Reasoning}: Covers ProntoQA \cite{saparov2023language}, LogiQA \cite{liu2020logiqa}, and ReClor \cite{yu2020reclor}, focusing on formal logic, deductive inference, and reasoning over structured premises.
    
    \item \textbf{Symbolic Reasoning}: CoinFlip \cite{wei2022chain} measures symbolic manipulation and stepwise logical computation.
    
    \item \textbf{Commonsense Reasoning}: Includes CommonsenseQA \cite{talmor2019commonsenseqa}, OpenBookQA \cite{OpenBookQA2018}, ECQA \cite{aggarwaletal2021ecqa}, and StrategyQA \cite{geva2021strategyqa}, assessing real-world plausibility, everyday knowledge, and context-aware implicit fact reasoning.
    
    \item \textbf{General Reasoning}: BIG-Bench \cite{srivastava2022beyond}, BIG-Bench Hard \cite{suzgun2022challenging}, HotPotQA \cite{yang2018hotpotqa}, MuSiQue \cite{trivedi2022musique}, MMLU \cite{hendrycks2020measuring}, MMMLU \cite{yue2024mmmu}, ScienceQA \cite{lu2022learn}, and SciBench \cite{wang2024scibench} jointly measure broad multi-domain reasoning, including complext multi-hop retrieval, factual synthesis, and robust interdisciplinary problem solving.
    
    \item \textbf{Visual Reasoning}: MMMU \cite{yue2024mmmu}, MATH-Vision \cite{wang2024measuring}, and MathVista \cite{lu2024mathvista} assess integration of visual perception with textual and mathematical reasoning.
    
    \item \textbf{Agent Reasoning}: TAU-bench~\cite{yao2024tau} and Keys-Finding Maze~\cite{su2025token} evaluate autonomous decision-making, planning, and environment interaction capabilities.
    
    \item \textbf{Task-specific Reasoning}: PubMedQA \cite{jin2019pubmedqa} measures biomedical question answering using domain-specific scientific literature, particularly focusing on reasoning.
\end{enumerate}



\begin{table}[!t]
\centering
\setlength{\tabcolsep}{3pt}
\caption{\textbf{Acc. and Token Costs of Implicit Latent CoT Methods.}}
\label{tab:benchmark_3}
\resizebox{1.0\linewidth}{!}{
\begin{tabular}{lccccc}
\toprule[1.1pt]
\textbf{Types} & \textbf{Methods}  & \textbf{Setting} & \textbf{Accuracy} & \textbf{Models} & \textbf{Token Costs} \\
\midrule
\multirow{10}{*}{\shortstack[l]{Implicit \\ Latent \\CoT}} &

ICoT-KD~\cite{deng2023implicit}  & zero-shot & 45.00\%  & GPT-2 Medium & -\\ 
& CODI~\cite{CODI}  & zero-shot & 55.60\% & LLaMA-3.2-1B & -\\
& ICoT-SI~\cite{deng2024explicit}  & zero-shot & 51.00\% & Mistral 7B & -\\
& COCONUT~\cite{hao2024training}  & zero-shot & 34.10\% & GPT-2 & 8.20\\
& CCoT~\cite{cheng2024compressed}  & zero-shot & 31.50\% & LLaMA2-7B-Chat & -\\
& Token assorted \cite{su2025token} & zero-shot & 37.20\% & LLaMA-3.1-8B &
-\\
& SoftCoT \cite{xu2025softcot}  & zero-shot & 85.81\% &
Qwen2.5-7B-Instruct &
- \\
& Efficient Latent Refinement \cite{wang2025efficient} & zero-shot & 40.20\% & GPT-2 & - \\
& \multirow{3}{*}{Soft Thinking \cite{zhang2025softthinking}} & zero-shot & 96.81\% & QwQ-32B & 1391 \\
&  & zero-shot & 95.83\% & DeepSeek-R1-Distill-Qwen-32B & 785 \\
&  & zero-shot & 94.90\% & DeepSeek-R1-Distill-LLaMA-70B & 597 \\

\bottomrule[1.1pt]
\end{tabular}}
\end{table}

In addition to categorizing benchmark tasks by reasoning scenario, we further provide a taxonomy of specific benchmark datasets used by the surveyed methods. Tables~\ref{tab:benchmark_explicit_1} and \ref{tab:benchmark_implicit} summarize which datasets are used by different efficient inference methods, grouped into explicit compact CoT and implicit latent CoT, respectively. This benchmark-level mapping enables a clearer view of method applicability across diverse reasoning settings.

\begin{table*}[!t]
\centering
\setlength{\tabcolsep}{3pt}
\captionsetup{skip=2pt}
\caption{\textbf{Analyses on Mathematical Objective Functions in Efficient Reasoning Methods (Part I)}}
\label{tab:rl_objectives_1}
\resizebox{1.0\linewidth}{!}{
\begin{tabular}{ccc}
\toprule[1.1pt]
\textbf{Name} & \textbf{Method} & \textbf{Objective Function} \\
\midrule
Budget Guidance \cite{li2025budget} & Inference-Time Guidance &
$\begin{array}{c}
p(Y_t \mid X, Y_{<t}, L_t \leq \bar{l} - t) \propto p(Y_t \mid X, Y_{<t}) \cdot \Pr(L_t \leq \bar{l} - t \mid X, Y_{<t}, Y_t) \\
c_t = \text{normalize}(u_t \circ a_t) \\
p(L_t \mid X, Y_{<t}, Y_t = v_i) = \text{Gamma}(\log(L_t); \lambda_t(v_i), \alpha_t(v_i))
\end{array}$\\
\rowcolor{gray!10} Overclocking LLM Reasoning \cite{eisenstadt2025overclocking} & Intervention & 
$\begin{array}{c}
\theta^* = \arg\min_\theta \sum_{(h,p)\in D} (f_\theta(h) - p)^2 \\
\hat{p} = \theta^T h \\
h_\alpha = h + \alpha \theta \\
\theta^T h_\alpha = \hat{p} + \alpha \|\theta\|^2
\end{array}
$ \\
Meta-Reasoner \cite{meta_reasoner} & Prompt & $
s_t = \arg\max_{s \in \mathcal{S}} \left( \mathbf{x}_t^\top \hat{\boldsymbol{\theta}}_s + c \sqrt{ \mathbf{x}_t^\top \mathbf{A}_s^{-1} \mathbf{x}_t } \right)
$
\\
\rowcolor{gray!10} AnytimeReasoner \cite{Qi2025OptimizingAR} & RL & $J_{\text{anytime}}(\theta, \phi) = \mathbb{E}_{x, z} \left[ \sum_{j=1}^{m} P_j \, r_\phi(x, z_{\leq b_j}) \right]$ \\
DuP-PO \cite{ding2025thinking} & RL &
\multicolumn{1}{>{\centering\arraybackslash}p{14cm}}{
\makecell[c]{$
\begin{array}{c}
\mathcal{J}(\theta) = \mathbb{E}_{D,(\pi_n,\pi_r)} \left[
\frac{1}{\sum_{i=1}^{N+M} |\tau_i|} \sum_{i=1}^{N+M} \sum_{t=1}^{|\tau_i|}
\min\left( \hat{r}_t^i \hat{A}_t^i,\; \text{clip}(\hat{r}_t^i, 1 - \epsilon, 1 + \epsilon) \hat{A}_t^i \right)
\right] \\
\quad - \beta \cdot D_{\mathrm{KL}}[\pi_\theta \parallel \pi_{\text{ref}}] \\
\hat{A}_t^i = m_t^i \cdot A_t^i \\
m_t^i =
\begin{cases}
\alpha, & \text{if } A_t^i > 0 \text{ and } \tau_i \sim \pi_r \\
\beta, & \text{if } A_t^i < 0 \text{ and } \tau_i \sim \pi_n \text{ and } \tau_{i,t} \in S_{\text{think}} \\
0, & \text{if } A_t^i > 0 \text{ and } \tau_i \sim \pi_n \text{ and } \tau_{i,t} \in S_{\text{think}} \\
1, & \text{otherwise}
\end{cases}
\end{array}$ }}\\
\rowcolor{gray!10} IBPO \cite{yu2025think} & RL & 
\multicolumn{1}{>{\centering\arraybackslash}p{14cm}}{
\makecell[c]{$
\begin{array}{c}
\hat{\pi}^*_{X,Y_\theta} \in \arg\max\limits_\pi \quad \\ \hat{J}_\Delta(\pi; X, Y_\theta) := \frac{1}{nm} \sum_{i=1}^{n} \sum_{j=1}^{m} \left[ \pi(y_{ij}|x_i) \cdot r_\Delta(x_i, y_{ij}) \right] \\
\text{s.t.} \quad  \pi \in \Pi \cap \hat{\Phi}^+(X, Y_\theta), \\
 \sum_y \pi(y|x) \cdot \mathbf{1}\{y \in \Xi_x\} \geq 1, \quad \forall x \in X
\end{array}
$}} \\
REO-RL \cite{gao2025how} & RL & 
$\begin{array}{c}
\mathcal{L}_{\text{Efficiency}}(\theta, D) = \sum_{L=1}^{L_{\text{max}}} J(D, \theta, L) \\
\mathcal{L}_{\text{REO-RL}}(\theta, D) = \mathbb{E}_{x \sim D} \left[ \mathbb{E}_{y \sim \pi_\theta(\cdot|x)} \left[ \sum_{i=1}^{N+1} c_i \cdot r(x, y{:}L_i; \theta) \right] \right] \\
d_{\text{REG}}(\theta, D, \hat{\Theta}) = \sum_{L=1}^{L_{\text{max}}} \left( \hat{J}_{\text{optimal}}(D, \hat{\Theta}, L) - J(D, \theta, L) \right)
\end{array}
$ \\
\rowcolor{gray!10} DART \cite{jiang2025dart} & SFT & 
$\begin{array}{c}
\mathcal{L}_{\text{DART}} = \mathcal{L}_{\text{CoT}} + \mathcal{L}_{\text{ST}} + \lambda \mathcal{L}_{\text{distill}} \\
\mathcal{L}_{\text{CoT}} = -\frac{1}{N} \sum_{i=1}^{N} \log p(z_i \mid Q, z_{1:i-1}; \theta) - \frac{1}{M} \sum_{i=1}^{M} \log p(y_i \mid Q, Z, y_{1:i-1}; \theta) \\
\mathcal{L}_{\text{ST}} = -\frac{1}{M} \sum_{i=1}^{M} \log p(y_i \mid Q, X, y_{1:i-1}; \theta, \phi) \\
\mathcal{L}_{\text{distill}} = \frac{1}{L} \sum_{l=1}^{L} \frac{1}{\sigma(\tilde{h}^l)} \left\| \tilde{h}^l - \hat{h}^l \right\|_1
\end{array}$ \\
Prune-on-Logic \cite{zhao2025can} & SFT & 
\multicolumn{1}{>{\centering\arraybackslash}p{14cm}}{
\makecell[c]{
$\begin{cases} \text{Score}_i = \text{PPL}_{\text{prune}} - \text{PPL}_{\text{retain}}\\ \text{PPL}_{\text{retain}} = \exp\left( \frac{1}{L_i} \sum_{j=p_s}^{p_e} \sum_{k=1}^{t_j} -\log P\left(\text{tok}_j^k \mid s_{<j}, \{\text{tok}_j^l\}_{l<k}; \text{SLM} \right) \right)\\ \text{PPL}_{\text{prune}} = \exp\left( \frac{1}{L_i} \sum_{j=p_s}^{p_e} \sum_{k=1}^{t_j} -\log P\left(\text{tok}_j^k \mid s_{<j} \setminus n_i, \{\text{tok}_j^l\}_{l<k}; \text{SLM} \right) \right) \end{cases}$ }} \\
\rowcolor{gray!10} AdapThink \cite{wan2025adapthink} & SFT, RL & 
\multicolumn{1}{>{\centering\arraybackslash}p{14cm}}{
\makecell[c]{
$\begin{array}{c}
\mathcal{L}_{\text{total}} = \mathcal{L}_{\text{GRPR}} + \mathcal{L}_{\text{accuracy}} \\
r(x, G, \theta) = \text{clip}\left( \left| \omega(\varphi) \right| \cdot (\lambda_o - \lambda_l) + \mathbb{I}[\omega(\varphi) < 0] \cdot \omega(\varphi) \cdot \lambda_b, \, r_{\text{min}}, \, r_{\text{max}} \right) \\ 
\omega(\varphi) = 
\begin{cases}
+1 & \text{if } \varphi \leq \varphi_{\text{low}} \\
\cos\left( \frac{\varphi - \varphi_{\text{low}}}{\varphi_{\text{high}} - \varphi_{\text{low}}} \cdot \pi \right) & \text{if } \varphi_{\text{low}} < \varphi < \varphi_{\text{high}} \\
-1 & \text{if } \varphi \geq \varphi_{\text{high}}
\end{cases} 
 \end{array}$ }}\\
LC-R1 \cite{cheng2025optimizing} & SFT, RL & 
$\begin{array}{c}
\mathcal{L}_{\text{total}} = \mathbb{E}_{q, \{o_i\}} \left[
\frac{1}{\sum_{i=1}^{G} |o'_i|} \sum_{i=1}^{G} \sum_{t=1}^{|o'_i|} 
\min\left(R_t(\theta) \cdot \hat{A}_{i,t}, \text{clip}(R_t(\theta), 1 - \epsilon, 1 + \epsilon) \cdot \hat{A}_{i,t}\right)
- \beta \cdot \text{KL}[\pi_\theta \| \pi_{\text{ref}}]
\right] \\
\hat{A}_{i,t} = r_{i,\text{combine}} + \gamma \cdot \mathbb{I}(o'_{i,t} = \texttt{</think>}) \cdot r_{i,\text{compress}} \\
r_{i,\text{length}} = 1 - \frac{|o'_i|}{\max_{j \in C} |o'_j|} \\
\makecell[c]{
r_{i,\text{compress}} = \begin{cases}
1 - \frac{|t(o'_i)|}{|t(o_i)|}, & \text{if correct and answer in } t(o'_i) \\
-1, & \text{if correct and answer not in } t(o'_i) \\
0, & \text{if wrong}
\end{cases}}
\end{array}$\\
\rowcolor{gray!10} DTO \cite{an2025think} & SimPO & \makecell[c]{$
\begin{aligned}
\text{minimize} \quad & \mathbb{E}_{x \sim D} \left[ C(\Delta_x) \right] \\
\text{subject to} \quad & \mathbb{E}_{x \sim D} \left[ P(\Delta_x) \right] \geq \alpha
\end{aligned}
$}
\\
\bottomrule[1.1pt]
\end{tabular}}
\end{table*}

\subsection{Analyses on Performance \& Efficiency}
In this subsection, we present a comprehensive examination of the performance and token consumption associated with a set of existing methods when applied to the widely used GSM8K dataset \cite{cobbe2021gsm8k}. The evaluation incorporates multiple methods, multiple models, and multiple experimental settings, ensuring that the comparison reflects a broad spectrum of approaches under diverse configurations.
\begin{enumerate}[label=\textbullet, leftmargin=0.4cm, itemsep=0.2em, parsep=0.2em, topsep=0.em]  
   \item \textbf{GPT-4o} is widely regarded for strong performance in complex reasoning and multi-turn problem solving, consistently ranking among the top models on publicly reported benchmarks. Its large parameter scale and extensive training enable high accuracy, but this also leads to increased computational requirements, longer inference times, and higher token usage per query. 
   
   \item \textbf{LLaMA-3.1} delivers strong competitive reasoning performance across diverse benchmarks while generally offering lower inference cost than larger proprietary systems. Its open-weight availability facilitates reproducibility and experimentation, though performance may be slightly lower in highly specialized reasoning tasks. 
   
   \item \textbf{Claude 3.5 Sonnet} provides balanced performance, demonstrating robust results in reasoning, summarization, and long-context processing. It is recognized for maintaining efficiency when handling extended inputs, keeping inference latency and token usage moderate relative to output quality, especially in practice. 
   
   \item \textbf{DeepSeek-R1} targets reasoning-intensive applications and maintains stable performance on tasks requiring structured step-by-step outputs. While its overall computational cost is moderate, token efficiency depends on the complexity of the reasoning chain generated. 
   
   \item \textbf{QwQ-32B} achieves strong results for its parameter size, with broad coverage of general knowledge and reasoning capabilities. However, it incurs higher per-query computational cost compared to smaller models, making it less optimal for resource-limited deployments. 
   
   \item \textbf{Distilled models} such as R1-Distill-LLaMA, DeepSeek-R1-Distill-Qwen, and DeepSeek-R1-Distill-LLaMA consistently retain a substantial portion of their teacher models’ reasoning accuracy while significantly lowering latency and reducing token consumption, making them well-suited for environments with constrained compute or strict response-time requirements. 
   
   \item \textbf{Qwen-2.5-Math} is oriented toward mathematical and symbolic reasoning tasks, showing consistent accuracy in domain-specific benchmarks. Its outputs tend to be concise, which can improve token efficiency in scenarios where step-by-step elaboration is not essential. 
   
   \item \textbf{Falcon-40B} demonstrates strong general-purpose capabilities relative to its scale, effectively addressing a wide range of reasoning and comprehension tasks; however, it lags behind models extensively fine-tuned for multi-step, complex reasoning in terms of peak accuracy.
\end{enumerate}



The analysis focuses on quantitatively comparing the accuracy achieved by each method alongside the corresponding token costs incurred during inference. All experimental results are systematically organized and reported in Table~\ref{tab:benchmark_1} and Table~\ref{tab:benchmark_3}, thereby providing a clear and structured basis for subsequent comparative analyses.

\subsection{Analyses on Objective Functions}

To complement the proposed categorization of efficient reasoning paradigms, we conduct a systematic analysis of objectives adopted in representative methods. The corresponding mathematical formulations are summarized in Table~\ref{tab:rl_objectives_1}, and Table 9 and Table 10 of the Appendix, spanning from prompting-based strategies to supervised fine-tuning (SFT), preference optimization, reinforcement learning (RL), self-training frameworks (STF), and inference-time intervention.
\begin{enumerate}[label=\textbullet, leftmargin=0.4cm, itemsep=0.2em, parsep=0.2em, topsep=0.em]  
   \item \textbf{Prompt-based methods} influence model outputs without parameter updates by modifying decoding scores through token-level likelihoods, structural constraints, or logical validity checks, especially during inference.
   
   \item \textbf{SFT-based approaches} minimize the cross-entropy loss over curated reasoning trajectories, aligning the model distribution $P_\theta(y|x)$ with human-verified solutions.
   
   \item {Preference optimization methods}, such as Direct Preference Optimization (DPO) and SimPO, extend SFT by introducing pairwise ranking losses. DPO directly optimizes for the likelihood ratio between preferred and dispreferred outputs, while SimPO incorporates similarity constraints to stabilize optimization, encouraging outputs that are both preferred and semantically consistent.
   
   \item \textbf{RL-oriented approaches} define explicit reward functions $R(y, x)$ encoding correctness, stepwise validity, or efficiency constraints. Policy gradient algorithms maximize expected rewards, enabling alignment with non-differentiable evaluation metrics.
   
   \item \textbf{Self-Training Frameworks (STF)} iteratively generate pseudo-labels for unlabeled data and optimize likelihood-based objectives over these augmented datasets. This reduces reliance on costly annotations while propagating reasoning patterns discovered during inference, effectively enhancing generalization.
   
   \item \textbf{Inference-time intervention methods} preserve model parameters and instead manipulate the decoding process through differentiable scoring terms, constraint-satisfaction formulations, or dynamic search strategies. This allows task-specific adaptation without retraining.
\end{enumerate}

These objectives reflect the optimization principles underpinning the reasoning capabilities of modern LLMs.

\section{Limitations \& Challenges}

\begin{figure}[!t]
\centering
\small
\begin{minipage}{1.0\linewidth}
\centerline{\includegraphics[width=1\textwidth]{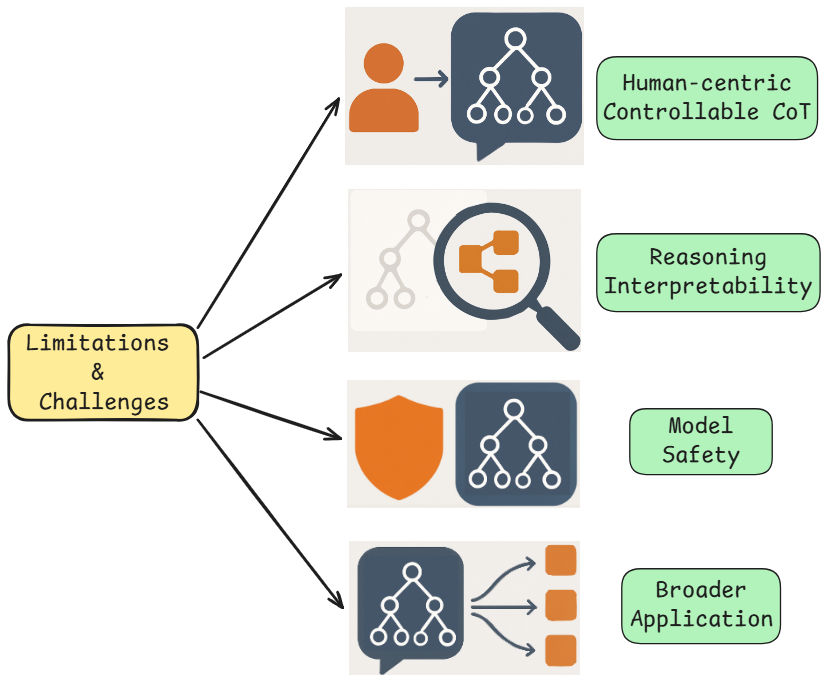}}
\end{minipage}
\caption{\textbf{Limitations and Challenges in Reasoning Efficiency.} The image highlights key challenges such as Human-centric Controllable CoT, Reasoning Interpretability, Model Safety, and Broader Application.}
\label{fig:limitations_and_challenges}
\end{figure}

Although the recent efficient reasoning methods have achieved promising performance, there are still several important limitations that hinder their widespread adoption and full effectiveness. 
To this end, we discuss the limitations and challenges of the existing efficient reasoning methods from the perspectives of \textbf{user experience}, \textbf{interpretability}, \textbf{safety}, and \textbf{application}, as shown in Figure~\ref{fig:limitations_and_challenges}.

\subsection{User-centric Controllable Reasoning}
Recent advancements in LRMs, such as OpenAI's o3~\cite{openai2025systemcard} and Anthropic's Claude 3.7~\cite{claud3_7}, have introduced \textbf{user-configurable reasoning modes}, allowing users to choose whether the model engages in explicit reasoning or provides direct answers. Additionally, these models enable users to control the complexity and length of the reasoning process, adapting to different needs and preferences.

This level of control is especially useful in diverse applications, e.g., in \textbf{educational settings}, users may prefer detailed step-by-step explanations for questions, whereas in \textbf{real-time decision-making tasks}, concise responses are typically more desirable. The ability to allow users to adjust reasoning depth enables LRMs to effectively balance efficiency and transparency, thereby enhancing user experience.

Future research should explore more refined control mechanisms, such as \textbf{interactive reasoning settings} that dynamically adjust based on user feedback. Besides, building personalized reasoning profiles could allow LRMs to learn and adapt to user preferences over time, providing a balance between reasoning depth, speed, and interpretability.

\subsection{Trade-off Between Interpretability and Efficiency}
Compared to LLMs, LRMs offer better \textbf{interpretability} due to their structured reasoning process. By explicitly generating intermediate reasoning steps, LRMs allow users to trace how a conclusion is reached, making them particularly valuable for applications where \textbf{transparency} and verifiability are critical, such as \textbf{scientific research}~\cite{rane2023contribution}, \textbf{medical diagnosis}~\cite{ullah2024challenges}, and \textbf{legal decision-making}~\cite{cheong2024not}. 
However, current efficiency-focused LRMs may compromise this interpretability. Many recent methods designed to accelerate LRM inference reduce the number of explicit reasoning steps or shift reasoning to latent representations, making it harder to understand how a model arrives at its conclusions.

Also, the importance of interpretability varies depending on the application. In domains such as healthcare and legal reasoning, where explanations are essential for accountability and human oversight, explicit reasoning steps are preferred despite their computational cost. Conversely, in real-time decision-making tasks, such as automated trading or robotics, efficiency often takes precedence over transparency, making implicit reasoning more desirable. Hybrid approaches, which dynamically adjust the level of explicit reasoning based on task complexity, offer a potential solution but require further refinement to prevent critical reasoning steps from being lost in the pursuit of efficiency.

To address this trade-off more effectively, future research should focus on developing adaptive inference strategies that optimize the balance between reasoning efficiency and interpretability. One promising direction is the integration of \textbf{external verification mechanisms}, such as symbolic reasoning~\cite{besold2021neural,gaur2023reasoning,sui2024fidelis,sui2024can} or retrieval-based justifications~\cite{gao2023retrieval}, which can provide post-hoc explanations for implicit reasoning models. Besides, new empirical studies are needed to systematically quantify how different efficiency techniques impact both model accuracy and human trust, guiding the development of LRMs that are both efficient and interpretable in real-world scenarios.

\subsection{Ensuring Safety of Efficient Reasoning}
Although the existing methods improve the token efficiency of the LRMs, they may destroy the alignment of LRMs, increasing the potential safety risks, e.g., \textbf{jailbreaking attacks} \cite{liuyue_FlipAttack,he2025evaluating} and \textbf{privacy leakage} \cite{privacy_attack,wang2025safety}. 

Firstly, the current training-based token-efficient methods either train the LRMs to prefer shorter generations \cite{kang2024c3ot, TALE} or adopt RL and incentivize concise responses via rule-based reward \cite{qu2025optimizing,luo2025o1,team2025kimi}. Given that the safety alignment is conducted on the original long reasoning generations and the safety of the shorter reasoning generations can not be guaranteed, these training processes might \textbf{break the safety alignment} of the original LRMs. 

Secondly, as one piece of evidence, researchers \cite{misbehavior_in_cot} found that the frontier LRMs tend to exploit the loopholes once they get a chance. In addition, although they tried to use another LLM to monitor the intermediate CoT, penalizing their misbehavior can not effectively alleviate this problem but further guide them to deliberately \textbf{hide their misconduct intent}. From this phenomenon, we suspect that the existing token-efficient methods unintentionally guide the LRMs to hide their harmful intent during the process of making their response more concise, significantly increasing the difficulty of safeguarding LRMs. 

To address this problem, one promising direction is to strictly enforce \textbf{safety constraints} during the training process, like data filtering for the SFT/DPO data and designing the safety-related reward in RL training. Besides, the failure of current monitors may be due to LRMs' ability being stronger than LLM-based guard models. Thus, it is worth designing stronger \textbf{reasoning-based safeguard models} \cite{liuyue_GuardReasoner,liuyue_GuardReasoner-VL} to monitor the training data or LRMs.

\subsection{Broader Application of Efficient Reasoning}

As shown in Table \ref{tab:taxonomy_3_1}, \ref{tab:taxonomy_3_2}, \ref{tab:taxonomy_4}, existing LRMs are primarily applied in specialized domains including \textbf{math} \cite{xia2025tokenskip,li2025solar,gong2025efficient}, \textbf{code} \cite{team2025kimi}, and \textbf{AI research} \cite{deep_research} scenarios. 

The first reason is that these tasks have relatively fixed answers, making it easier to construct objectives, e.g., preparing reasoning data, formulating preference loss functions, or rule-based rewards. In contrast, other domains, like \textbf{social sciences} \cite{thapa2025large}, \textbf{emotional intelligence} \cite{wu2025personas}, \textbf{creative writing} \cite{gpt4_5}, typically involve open-ended questions, making it difficult to formulate clear objectives. 

The second reason is that these scenarios, like math or research, are not highly time-sensitive, allowing for more computational resources to be allocated for reasoning and optimization. The high computational demand and latency of LRMs constrain their applicability in broader time-sensitive domains, such as \textbf{robotic manipulations} \cite{ji2025robobrain,geimini_vla,nvidia_n1}, \textbf{financial trading} \cite{ding2024large}, \textbf{autonomous driving} \cite{yang2023llm4drive}. 

However, recently developed efficient reasoning methods \cite{cheng2024compressed,claud3_7,team2025kimi} help LRMs reduce thinking tokens, optimize timing and memory usage, and thus \textbf{enhance feasibility in real-time applications}. For the open-ended questions, efficient reasoning methods enable LRMs to generate more structured and consistent responses while \textbf{balancing interpretability and computational cost}.

\subsection{Takeaways of Limitations Challenges}
Despite the advantages of efficient reasoning methods, they also pose several challenges. The following summarizes key limitations and potential directions for addressing them.

\begin{tcolorbox}[colframe=black, colback=gray!5, coltitle=white, breakable, title=Takeaways of Limitations Challenges]
\begin{itemize}[left=0pt, label=\textbullet, itemsep=1.2em]

    \item User-controllable reasoning enables users to adjust reasoning depth, striking a balance between transparency and efficiency while optimizing user experience. Future research should focus on interactive and personalized reasoning for users.

    \item Efficient reasoning methods may obscure crucial reasoning processes, compromising the interpretability of LRMs. Future research should develop adaptive inference strategies to balance efficiency and interpretability.

    \item Efficient reasoning methods may compromise safety alignment, increasing the risk of jailbreaking and privacy leakage. Future work should integrate safety constraints in training and develop stronger reasoning-based safeguards.

    \item Efficient reasoning methods may improve the feasibility of LRMs for broader applications like real-time applications and open-ended tasks.

\end{itemize}
\end{tcolorbox}

\section{Further Improvement}

\begin{figure}[!t]
\centering
\small
\begin{minipage}{1.0\linewidth}
\centerline{\includegraphics[width=1\textwidth]{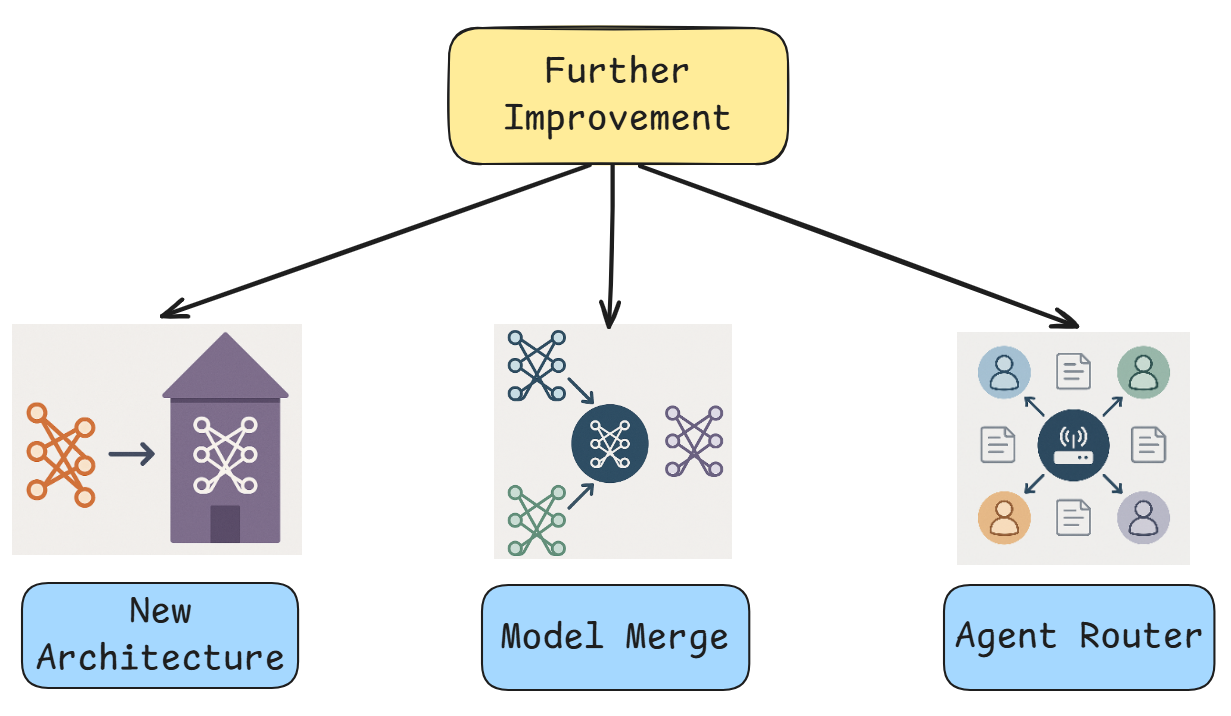}}
\end{minipage}
\caption{\textbf{Further Improvement Strategies for Reasoning Efficiency.} Directions include new architectures, model merge, and agent router.}
\label{fig:further_improvement}
\end{figure}

Although current approaches achieve strong performance, we present alternative strategies that could further improve inference efficiency while maintaining high reasoning quality, as shown in Figure~\ref{fig:further_improvement}, including \textbf{new architectures}, \textbf{model merge}, and \textbf{agent router}.

\subsection{New Architecture}
\textbf{Hybrid Autoregressive and Diffusion Models.} The fundamental limitation of autoregressive models is their sequential nature, which makes inference slow, particularly for reasoning tasks that require long chains of intermediate steps. A potential solution is integrating \textbf{diffusion models} into LRMs~\cite{llada}. 
Diffusion models generate entire sequences in parallel, allowing for global reasoning structure optimization rather than token-by-token generation. 
However, the challenge lies in controlling the generated reasoning steps to ensure logical consistency. 
A promising direction is hybrid architectures that use autoregression for fine-grained control over reasoning while leveraging \textbf{diffusion-based sampling} for efficiency, enabling LRMs to reason in a structured yet accelerated manner. 
While diffusion offers potential speedups, the overhead of managing this synchronicity and the potential need for multiple iterative refinement to correct logical inconsistencies might offset some gains, especially when compared to optimized autoregressive approaches like speculative decoding, which also aim to accelerate generation without sacrificing as much direct control. The actual trade-off between generation speed, resource consumption (as diffusion models can be computationally intensive to train and sample from), and the quality of reasoning remains an open research question.

\textbf{Memory-Efficient Transformer Variants.} One of the primary inefficiencies in LRMs stems from the \textbf{quadratic complexity} of self-attention. Applying \textbf{linear attention} mechanisms(e.g., RWKV~\cite{peng2023rwkv}) or \textbf{state-space} models (e.g., Mamba~\cite{gu2023mamba}) could drastically reduce memory consumption and improve inference speed. The challenge is that such architectures often struggle with \textbf{long-range dependencies}, which are crucial for reasoning. A key question is whether \textbf{hybrid models} can selectively apply full attention for critical reasoning steps while using approximate attention elsewhere to optimize efficiency. 
The practicality of such hybrids depends on the effective identification of "critical" steps and the seamless integration of different attention mechanisms without introducing excessive architectural complexity or training instability. Compared to methods like quantization or sparse attention, which aim to reduce the cost of full attention, linear attention and state-space models represent a more fundamental architectural shift. The actual benefit will depend on whether the reduced memory and computation translate to tangible improvements in reasoning quality per unit of resource, especially for tasks demanding high fidelity.


\subsection{Model Merge}

The underlying principle of the existing token-efficient methods can be summarized as integrating the strength of the conventional LLMs, i.e., \textbf{fast responses and low costs}, and the strength of the LRMs, i.e., \textbf{deliberative reasoning and accurate responses}.

The existing training-based methods \cite{TALE,luo2025o1,qu2025optimizing} typically involve reasoning data curation and post-training techniques such as SFT, DPO, or RL, making the process \textbf{complex and expensive}. In contrast, the existing training-free methods \cite{meta_reasoner} typically just use promoting engineering to guide the LRMs to save the tokens, \textbf{limiting the adaptability and effectiveness} across diverse reasoning tasks.

To solve this problem, another training-free method model merge \cite{yang2024model,wu2025unlocking} becomes a promising technique. Concretely, we can simply \textbf{merge the model weights} of one conventional LLM and the corresponding LRM to take their advantages together \cite{team2025kimi}. During this process, we provide several key points that need to be solved in the future. First, we need to \textbf{determine which modules or neurons in models should be merged}. Should we merge the neurons in shallow networks or deep networks? Then, we should \textbf{assign merging weights for the merging units}. Should we assign static or dynamic weights for each unit? Third, we should consider \textbf{how to merge models with different architectures and model sizes}, e.g., LLaMA-3.1 Instruct 8B \cite{llama_3} and DeepSeekR1-Distill-Qwen-7B \cite{guo2025deepseek}.

Merging models with disparate architectures or significantly different layer counts presents a particularly substantial technical hurdle. Simple averaging is often not viable. This may require more sophisticated techniques like parameter subspace alignment, or focusing the merge on specific, architecturally compatible layers (e.g., only attention or feed-forward layers if they share dimensions), which might limit the extent of synergistic merging. Compared to other ensemble methods (like distillation or mixture-of-experts where models are used more distinctly), model merging aims for a deeper integration at the parameter level. However, its practical advantage over simpler ensembling or parameter-efficient fine-tuning (PEFT) methods applied to a single base model needs to be clearly demonstrated, especially concerning the development effort and the consistency of performance gains.

\subsection{Agent Router}
Agent routing could further improve efficiency by directing different parts of a query to specialized agents. By routing the \textbf{query to the most appropriate agent} based on task complexity, this strategy would optimize resource usage and enable faster inference, particularly for tasks that require domain-specific knowledge or specialized reasoning.

Two routing strategies for LLM inference are router models and confidence-based metrics. Router models (e.g., Routellm \cite{ong2025routellm}) select between stronger or weaker LLMs to balance cost and quality. Confidence-based routing (e.g., Self-REF \cite{chuang2025learning}) directs queries based on LLMs' confidence in their answers, while uncertainty-based routing (e.g., SLM routing \cite{chuang2025confident}) offloads high-stakes queries to more robust models when confidence is low.

These approaches improve inference efficiency by reducing computation and resource usage while maintaining performance. However, agent routing, though promising, introduces challenges like system complexity, model version management, and operational costs. It only delivers a clear advantage when specialization and accurate routing lead to significant improvements over simpler strategies like multi-pass inference with a single scalable model.

\subsection{Takeaways of Further Improvement}
We summarize key future directions for further improvement of efficient reasoning methods as follows.

\begin{tcolorbox}[colframe=black, colback=gray!5, coltitle=white, breakable, title=Takeaways of Further Improvement]
\begin{itemize}[left=0pt, label=\textbullet, itemsep=1.2em]
    \item New architectures like autoregressive-diffusion models and memory-efficient transformers hold promise, but managing logical consistency and long-range dependencies remains challenging.

    \item Model merging shows promise in combining LLM and LRM strengths, but challenges in module selection, merging weights, and handling architectural differences need further exploration.

    \item Agent routing offers potential for efficiency by directing queries to specialized agents, but its practical advantages over simpler strategies and the complexity of maintaining multiple models and routers need to be carefully evaluated.

\end{itemize}
\end{tcolorbox}

\section{Final Remarks}
This survey provides an overview of efficient inference techniques for large reasoning models, highlighting the challenges and recent advancements in this area. As reasoning models continue to grow in scale, the computational cost of inference becomes a major bottleneck, necessitating methods that improve efficiency while maintaining performance. We categorized existing approaches, discussing their trade-offs and practical implications. We hope this survey provides a foundation for further research in this area, encouraging the development of more effective and computationally feasible reasoning models.

\bibliography{0_arxiv}
\bibliographystyle{IEEEtran}

\clearpage

\onecolumn
\appendix

We analyze mathematical objective functions in efficient reasoning methods in Table \ref{tab:rl_objectives_2} and Table \ref{tab:rl_objectives_3}.

\begin{table*}[h]
\centering
\setlength{\tabcolsep}{3pt}
\captionsetup{skip=2pt}
\caption{\textbf{Analyses on Mathematical Objective Functions in Efficient Reasoning Methods (Part II)}}
\label{tab:rl_objectives_2}
\resizebox{1.0\linewidth}{!}{
\begin{tabular}{ccc}
\toprule[1.1pt]
\textbf{Name} & \textbf{Method} & \textbf{Objective Function} \\
\midrule
Soft Thinking \cite{zhang2025softthinking} & Decoding & $p(y \mid x) = \sum_{t_1, \ldots, t_m} \prod p(t_i \mid \cdot) \cdot p(y \mid x, t_{1:m})$ \\
\rowcolor{gray!10} NoWait \cite{wang2025wait} & Inference-Time Filtering & 
$\begin{array}{c}
\text{CoT} = \{ (\text{chunk}_i, a_i) \}_{i=1}^{n} \\
\text{Response} = (\text{CoT}, \text{summary}) \\
K_\alpha = \{ v \in V_\alpha \mid \exists k_s \in K, \text{ s.t. } \text{is\_substr}(k_s, v) \} 
\end{array}$ \\
Answer Convergence \cite{liu2025answer} & Inference-time &
$\begin{array}{c}
\mathcal{L} = -\frac{1}{T} \sum_{t=1}^{T} \left[ p_t \log \hat{p}_t + (1 - p_t) \log (1 - \hat{p}_t) \right] \\
 \hat{p}_t = \sigma(Wz_t + b) \\
y^*_{t} \leftarrow y^*_{t}  + \alpha \cdot \left( \max(y) - \frac{1}{|y|} \sum_i y_i \right) 
\end{array}
$ \\
\rowcolor{gray!10} ASC \cite{azizi2025activation} & Inference-time & 
$\begin{array}{c}
v^\ell = \frac{1}{N} \sum_{i=1}^{N} \left( h^\ell(q_i \oplus s_i)[-1] - h^\ell(q_i \oplus l_i)[-1] \right) \\
h^\ell(x_i) \leftarrow h^\ell(x_i) + \gamma v^\ell \quad \forall i \in [1, \text{decoding steps}] \\
\text{KL}(\text{softmax}(z) \| \text{softmax}(\tilde{z})) \leq \varepsilon \\
\gamma_{\text{max}} = \max \left\{ 0, \left(1 - \frac{L \gamma_{\text{raw}}}{4a} \right) \gamma_{\text{raw}} \right\}
\end{array}$ \\
Fractured Sampling \cite{liao2025fractured} & Inference-time Scaling & $p_{\text{seg}} = 1 - \prod_{t=1}^{H}(1 - p_t)^m$ \\
\rowcolor{gray!10} TS \cite{zhang2025making} & Intervention & 
$r = \hat{r} - \zeta(L)$
\\
RPC \cite{song2025reasoning} & KV Cache Compression & $\text{Importance}(t) = \frac{1}{2w + 1} \cdot \frac{1}{RH} \sum_{i = -w}^{w} \sum_{r,h} \text{Attn}^\ell_h(q_r, k_{t+i})$ \\
\rowcolor{gray!10} CTS \cite{lin2025controlling} & None (Plug-and-play) &
$\begin{array}{c}
h^l \leftarrow h^l + \alpha \cdot v^l \\
d(x_t) = \frac{1}{|L|} \sum_{l \in L} \text{JSD}(p_N(\cdot|x_{<t}), p_l(\cdot|x_{<t})) \\
\text{threshold} = \mu_W + \lambda \cdot \sigma_W
\end{array}$ \\
Efficient Latent Refinement \cite{wang2025efficient} & Post-training (training-free) & 
$\begin{array}{c}
\textnormal{Residual Refinement:} \quad  h_t = \alpha \cdot h_{t-1} + (1 - \alpha) \cdot f(h_{t-1}) \\
\textnormal{Contrastive Update:} \quad  h_t^{\text{updated}} = h_t + \eta \cdot \nabla_{h_t} \left[ \text{MSE}(h_t, h_t^{\text{good}}) - \text{MSE}(h_t, h_t^{\text{bad}}) \right]
\end{array}$ \\
\rowcolor{gray!10} Constrained-CoT \cite{nayab2024concise} & Prompt & $\frac{1}{N} \sum_{i=1}^{N} \mathbf{1}(\Gamma(\hat{y}_i), y_i) \times p(\hat{y}_i)$ \\
EfficientXLang \cite{ahuja2025efficientxlang} & Prompt &
$\begin{array}{c}
\widehat{\text{Pass@}k}(l, n) = \frac{1}{m} \sum_{i=1}^{m} \left[ 1 - \frac{\binom{n - c(x_i, y_i)}{k}}{\binom{n}{k}} \right] \\
c(x_i, y_i) = \sum_{r \in R^{(n)}(x_i)} \mathbb{1}\left[ \text{LLM}(x_i, r) = y_i \land \text{LID}(r) = l \right]
\end{array}$ \\
\rowcolor{gray!10} PREMISE \cite{yu2025premise} & Prompt& 
$\begin{array}{c}
\mathcal{L}_{\text{total}} = \lambda \cdot \nabla_{\text{text}} \mathcal{L}_{\text{acc}} + (1 - \lambda) \cdot \nabla_{\text{text}} \mathcal{L}_{\text{len}}, \; \lambda \in [0,1] \\
\text{IO}(r, q) = \frac{L(r) - L^*(q)}{L(r)} \\
\text{IU}(r, q) = 1 - \frac{k^*(r, q)}{L(r)} 
\end{array}$\\
SoT \cite{aytes2025sketch} & Prompt & 
  $T(l_i, l_o, B) = \tilde{t}^P_B(l_i) + \sum_{k = l_i}^{l_i + l_o - 1} t^D_B(k)$\\
\rowcolor{gray!10} ThinkLess \cite{li2025thinkless} & Prompt & $p(x_{1:(M+N)} \mid q) = \left( \prod_{i=1}^{M} p(x_i^{\text{reason}}) \right) \left( \prod_{j=1}^{N} p(x_j^{\text{answer}}) \right)$ \\
TrimR \cite{lin2025trimr} & Prompt & $\min_{c(\cdot)} \text{Infer\_Cost}(y_{<t'}) \quad \text{s.t.} \; \text{Perf}(X, y_{<t'}) \geq \text{Perf}(X, y_{<t})$ \\
\rowcolor{gray!10} CTS \cite{yuan2025not} & STF & $\min_{\tilde{y}} \, \text{dist}(A, \tilde{A}) + \lambda \|\tilde{y}\|_0$ \\
DAST \cite{DAST} & SimPO & 
\multicolumn{1}{>{\centering\arraybackslash}p{14cm}}{
\makecell[c]{ 
$\begin{cases} \max\left(-0.5 \cdot \frac{L(y) - L_\text{budget}}{L_\text{budget}} + 0.5, 0.1\right), & \text{if } S(y) = 1 \\ \min\left(0.9 \cdot \frac{L(y) - L_\text{budget}}{L_\text{budget}} - 0.1, -0.1\right), & \text{if } S(y) = 0 \end{cases}$}} \\
\rowcolor{gray!10} BINGO \cite{liu2025bingo} & SFT, RL & 
\makecell[c]{
$\begin{array}{c}
\mathcal{R}_{\text{BINGO}}(y) =
\begin{cases}
\lambda_c \cdot r_{\text{is}}(y), & \text{if correct} \\
\lambda_{\text{is}}^w \cdot \left[ r_{\text{is}}(y) - 1 \right] + \min(0, r_s(y) - \lambda_s^w), & \text{if incorrect}
\end{cases} \\[1ex]
\mathcal{J}_{\text{BINGO}}(\theta) = \mathbb{E}_t \left[ \min \left( r_t(\theta)\hat{A}_t,\ \text{clip}(r_t(\theta), 1 - \epsilon, 1 + \epsilon) \hat{A}_t \right) \right]
\end{array}$ }\\
Causal \cite{yu2025causal} & SFT, RL & 
$\begin{array}{c}
\text{PNS}(S, s_t, q) := P(A_S = y, A_{S'} \ne y) \\
\text{PS}(S, q) = P(A_{\text{do}(S)} = y \mid A \ne y, \bar{S}, q) \\
\text{PN}(S, s_t, q) = P(A_{\text{do}(s_{<t}, \bar{s}_t, s'_{>t})} \ne y \mid A = y, S, q)
\end{array}$ \\
\rowcolor{gray!10} TALE-PT \cite{TALE} & SFT, DPO & $\begin{aligned}
\mathcal{L}_{\text{CE}}(\theta) &= -\frac{1}{N} \sum_{i=1}^{N} \sum_{t=1}^{T_i} \log P(y_{i,t} \mid y_{i,<t}, x_i) \\
\mathcal{L}_{\text{DPO}}(\theta) &= -\frac{1}{N} \sum_{i=1}^{N} \log P_\theta(y_i \succ y'_i)
\end{aligned}$$\begin{aligned}
\mathcal{L}_{\text{CE}}(\theta) &= -\frac{1}{N} \sum_{i=1}^{N} \sum_{t=1}^{T_i} \log P(y_{i,t} \mid y_{i,<t}, x_i) \\
\mathcal{L}_{\text{DPO}}(\theta) &= -\frac{1}{N} \sum_{i=1}^{N} \log P_\theta(y_i \succ y'_i)
\end{aligned}$
\\
ReCUT \cite{jin2025recut} & SFT, RL &
$\begin{array}{c}
y^*_t = \arg\max \left( r(Y^{l}_{[t]}),\ r(Y^{s}_{[t]}) \right) \\
\mathcal{L}_\text{DPO}(D) = -\mathbb{E}_{(q, Y^+, Y^-)\sim D} \left[
\log \sigma \left(
\beta \log \frac{M(Y^+|q)}{M_\text{ref}(Y^+|q)} -
\beta \log \frac{M(Y^-|q)}{M_\text{ref}(Y^-|q)}
\right)
\right] \\
\theta_\text{merge} = \theta_\text{acc} + \alpha \cdot \text{Top}_x(\theta_\text{len})
\end{array}$\\
\rowcolor{gray!10} SReF \cite{liu2025efficient} & SFT, RL & 
\multicolumn{1}{>{\centering\arraybackslash}p{14cm}}{
\makecell[c]{
$\mathcal{P}_{\text{token}}^{\text{adjusted}} = 
\begin{cases}
0, & \text{if } \mathcal{P}_{\text{token}} < \theta \text{ and token} \in \{\text{"wait"}, \text{"Wait"}\} \\
\mathcal{P}_{\text{token}}, & \text{otherwise}
\end{cases}$
}}\\
SmartThinker \cite{he2025smartthinker} & SFT, RL &
\makecell[c]{
$\begin{cases}
r_{i,j} =
\begin{cases}
(1 - k_1 \sigma(\tilde{l}_{i,j}))(1 - k_2 \sigma(\tilde{n}_i)), & \text{if } a_i = a \\
-e^{-\frac{\rho \cdot \tilde{d}_{i,j}'}{k_0}}, & \text{if } a_i \ne a
\end{cases}
\\
A_{i,j} = \sum_{n=0}^{k_{i} - j} \gamma^n \cdot \tilde{r}_{i,j+n}
\end{cases}$ }\\
\rowcolor{gray!10} TLDR \cite{li2025tldr} & SFT, RL & 
$\begin{array}{c}
\mathcal{L}(\theta , \alpha ) = \sum _{i=1}^2 \alpha _i \cdot \delta _i \\
\delta_i = \phi_{\text{sys-}i,\text{bound}}(x) - \phi_{\text{sys-}i,\theta}(x) \\
\lambda_{\text{sys-1}} = \max\left( \dfrac{ \phi_{\text{sys-1, bound}} - \phi_{\text{sys-1}, \theta_{\text{proxy}}} }{ \phi_{\text{sys-1}, \theta_s} - \phi_{\text{sys-1}, \theta_l} },\ 0 \right) \\
\lambda_{\text{sys-2}} = \max\left( \dfrac{ \phi_{\text{sys-2, bound}} - \phi_{\text{sys-2}, \theta_{\text{proxy}}} }{ \phi_{\text{sys-2}, \theta_l} - \phi_{\text{sys-2}, \theta_s} },\ 0 \right)
\end{array}
$ \\
ConCISE \cite{qiao2025concise} & SFT, SimPO & 
$s_{i+1} = \pi_\theta(S_i) = 
\left\{
\begin{array}{ll}
\text{ReflectionStep}, & c_i < t_i \\
\text{NextStep}, & \text{otherwise}
\end{array}
\right.$ \\
\rowcolor{gray!10} CoLE \cite{zhao2025efficient} & SFT, RL & 
$\begin{array}{c}
\mathcal{L}_{\text{total}} = \mathcal{L}_{\text{comp}} + \mathcal{L}_{\text{latent}} \\
r(y_j) = \lambda_1 \cdot \mathbb{I}(y_j = y_i^*) - \lambda_2 \cdot \max\left(0, \ell(y_j) - \ell_{\text{Min\_Correct}}\right)\\
v_l = \mu_l^{\text{efficient}} - \mu_l^{\text{verbose}}, \quad h'_l = h_l + \lambda v_l
\end{array}$\\
\bottomrule[1.1pt]
\end{tabular}}
\end{table*}

\begin{table*}[!t]
\centering
\setlength{\tabcolsep}{3pt}
\captionsetup{skip=2pt}
\caption{\textbf{Analyses on Mathematical Objective Functions in Efficient Reasoning Methods (Part III)}}
\label{tab:rl_objectives_3}
\resizebox{1.0\linewidth}{!}{
\begin{tabular}{ccc}
\toprule[1.1pt]
\textbf{Name} & \textbf{Method} & \textbf{Objective Function} \\
\midrule
FlashThink \cite{jiang2025flashthink} & Prompt, SFT & $y = \text{LLM}_\theta(x \mid r) = \text{LLM}_\theta(x \mid c_1, c_2, \dots, c_{|r|})$ \\
\rowcolor{gray!10} AALC \cite{li2025aalc} & RL &
$\begin{array}{c}
\mathcal{L}_{\text{AALC}} = \text{Att}_{\text{acc}} \cdot R_{\text{raw}} + \alpha \cdot R_{\text{len}} \\
r_{\text{acc}} = \frac{A_{\text{val}}}{A_{\text{target}}}, \quad
r_{\text{len}} = \min\left(1, \frac{L_{\text{pred}}}{L_{\text{max}}}\right) \\
R_{\text{len}} = 1 - \min\left(r_{\text{acc}}^{\beta}, r_{\text{len}}\right) \\
\text{Att}_{\text{acc}} = \gamma + (1 - \gamma)(1 - r_{\text{acc}})
\end{array}$ \\
ConciseR \cite{song2025walk} & RL & $J_{\text{GRPO++}}(\theta) = \mathbb{E} \left[ \min(\tau_i(\theta)\hat{A}_i, \text{clip}(\tau_i)) + \alpha \mathcal{H}(\pi_\theta) \right]$ \\
\rowcolor{gray!10} ConciseRL \cite{Dumitru2025ConciseRL} & RL & $J(\theta) = \mathbb{E}_{x \sim \rho} \mathbb{E}_{y \sim p_\theta(\cdot | x)} [ R(y, x) ]$ \\
Elastic Reasoning \cite{xu2025scalable} & RL & $J(\theta) = \mathbb{E}_{x \sim D,\, y \sim \pi_\theta(\cdot \mid x; t^*, s^*)} [ r(y) ]$ \\
\rowcolor{gray!10} ERL\cite{arora2025training} & RL & $\mathbb{E}_{x \sim \rho} \mathbb{E}_{y \sim p_\theta(x)} \left[ \mathbf{1}\{y = y^\star(x)\} \cdot \left(1 - \alpha f(\text{LEN}(y))\right) \right]$
\\
Kimi k1.5 \cite{team2025kimi} & RL & 
\multicolumn{1}{>{\centering\arraybackslash}p{14cm}}{
\makecell[c]{ 
$S(y) + \begin{cases} 0.5 \cdot \frac{L(y) - L_\text{min}}{L_\text{max} - L_\text{min}}, & \text{if } S(y) = 1 \\ \min(0, 0.5 - \frac{L(y) - L_\text{min}}{L_\text{max} - L_\text{min}}), & \text{if } S(y) = 0 \end{cases}$}} \\
\rowcolor{gray!10} L1 \cite{l1} & RL & 
\makecell[c]{$
r(y, y_{\text{gold}}, n_{\text{gold}}) =
\begin{cases}
1 - \alpha \cdot \left| n_{\text{gold}} - n_y \right|, & \text{if exact length constraint is used (L1-Exact)} \\
\mathbb{I}(y = y_{\text{gold}}) \cdot \text{clip}\left(\alpha \cdot (n_{\text{gold}} - n_y) + \delta,\, 0,\, 1\right), & \text{if max length constraint is used (L1-Max)}
\end{cases}
$}
\\
LASER \cite{liu2025learn} & RL & $\hat{R}(x, y) = C(y) + \lambda(y) \cdot S(y)$ \\
\rowcolor{gray!10} Length-Aware Optimization \cite{Danlong2025EfficientRL} & RL & 
\multicolumn{1}{>{\centering\arraybackslash}p{14cm}}{
\makecell[c]{
$\text{reward}_{\text{len}}(i) = \begin{cases} \beta, & r(x, y_i, y^*) > 0 \land \text{acc} \geq \text{acc}_{\max} - \tau_{\text{acc}} \\ 0, & \text{otherwise} \end{cases}$ }}\\
MRT \cite{qu2025optimizing} & RL & $\Delta^{\mu}_{k}(x; \pi) := \mathbb{E}_{z \sim \pi(\cdot | x)} \left[ \sum_{j=0}^{k-1} \left( J_r(x; \pi^*_j) - J_r(x; \mu(\cdot | x, z_{0:j})) \right) \right]$
\\
\rowcolor{gray!10} O1-Pruner \cite{luo2025o1} & RL & $\frac{L_\text{ref}}{L(y)} - 1 + \lambda(S(y) - S(y_\text{ref}))$ \\
S-GRPO \cite{dai2025sgrpo} & RL & $J_{\text{S-GRPO}}(\theta) = \mathbb{E}_{q, \{o_i\}} \left[ \frac{1}{G} \sum_{i=1}^{G} \frac{1}{|o_i|} \sum_{t=1}^{|o_i|} \left\{ \min \left[ \frac{\pi(o_{i,t})}{\pi_{\text{old}}(o_{i,t})} \hat{A}_{i,t}, \text{clip}(\cdot) \hat{A}_{i,t} \right] \right\} \right]$ \\
\rowcolor{gray!10} SelfBudgeter \cite{li2025selfbudgeter} & RL & 
\multicolumn{1}{>{\centering\arraybackslash}p{14cm}}{
\makecell[c]{
$R(C, F, \ell, b) = \begin{cases} r_f & F = 0 \\ PB + \text{PreB}(\cdot) & F = 1 \end{cases}$ }} \\
SPIRIT \cite{cui2025stepwise} & RL & $\text{PPL}(x, \{w_k\}_{k=1}^N) = \exp\left( -\frac{1}{N} \sum_{i=1}^N \log p(w_i \mid x, w_1, \ldots, w_{i-1}) \right)$
\\
\rowcolor{gray!10} TLDR \cite{zhang2025making} & RL & 
\multicolumn{1}{>{\centering\arraybackslash}p{14cm}}{
\makecell[c]{
$r = \hat{r} - \zeta(L), \quad \zeta(L) = \begin{cases} 0 & \text{(length within bounds)} \\ \beta & \text{length exceeded)} \\ \eta(L) & \text{otherwise} \end{cases}$ }}\\
A*-Thought \cite{xu2025a} & SFT & $f(t'_k + r_w) = g(t'_k + r_w) + h(t'_k + r_w)$ \\
\rowcolor{gray!10} Adaptive GoGI-Skip \cite{zhuang2025accelerating} & SFT & $G_t^{(l^*)} = \left\| \frac{\partial L_{\text{ans}}}{\partial h_t^{l^*}} \right\|_1$ \\
C3oT \cite{kang2024c3ot} & SFT &  $\{(x_i, r^{\text{long}}_i, y_i)\}_{i=1}^N$
\\
\rowcolor{gray!10} CCoT~\cite{cheng2024compressed} & SFT & 
$\text{LOSS}_{\varphi}(z_i^\ell, \hat{z}_i^\ell) = \frac{1}{k} \sum_{i=1}^{k} \frac{1}{\sigma^2(z_i^\ell)} \, \text{MSE}(z_i^\ell, \hat{z}_i^\ell)$
\\
COCONUT~\cite{hao2024training} & SFT & $H_t = \text{Transformer}(E_t) ;\mathcal{M}(x_{t+1} \mid x_{\leq t}) = \text{softmax}(W h_t)$\\
\rowcolor{gray!10} CODI~\cite{CODI} & SFT & $\mathcal{L} = \alpha \mathcal{L}_{\text{teacher}} + \beta \mathcal{L}_{\text{student}} + \gamma \mathcal{L}_{\text{KD}}$
\\
CoT-Valve \cite{ma2025cot} & SFT & $\begin{aligned}
&p(a \mid t_1, \ldots, t_n, q; \theta) \prod_{i=1}^{n} p(t_i \mid t_{<i}, q; \theta);  \max_{\Delta\theta} \; \mathbb{E}_{(q,a) \sim D} \left[
p(a \mid t_1, \ldots, t_m, q; \theta + \Delta\theta)
\prod_{i=1}^{m} p(t_i \mid t_{<i}, q; \theta + \Delta\theta)
\right] \\
\end{aligned}$\\
\rowcolor{gray!10} Distill System 2 \cite{yu2024distilling} & SFT & 
$S_{II}(x; p_\theta) \rightarrow z, y$
\\
DRP \cite{jiang2025drp} & SFT & $\mathcal{L}_{\text{SFT}} = - \sum_{i=1}^{n} \log P_{\theta}(y_i \mid x, y_{<i})$ \\
\rowcolor{gray!10} Heima \cite{shen2025efficient} & SFT & $P_{\theta} \left( \left\{ <CoT>^{(k)} \right\}_{k=1}^{K_i}, Y_a^{(i)} \mid X_v^{(i)}, X_q^{(i)} \right)
$\\
ICoT-KD~\cite{deng2023implicit} & SFT & $P(y | x) \approx \int_{\hat{z}} P_\theta(\hat{z} | x) P_\theta(y | x, \hat{z})$
\\
\rowcolor{gray!10} ICoT-SI~\cite{deng2024explicit} & SFT & $\min_{\theta} - \log P_{\theta}(y, z_{1:m} \mid x)$\\
InftyThink \cite{yan2025inftythink} & SFT &
\multicolumn{1}{>{\centering\arraybackslash}p{14cm}}{
\makecell[c]{$
\text{For } i = 1 \text{ to } n:\quad
\displaystyle
\begin{cases}
S_i = \text{summarize}(M, RP_i, \{RP_j\}_{j=1}^{i-1}) & \text{if } i < n \\
\text{Conclusion} = \text{generate}(M, RP_n, S_{n-1}) & \text{if } i = n
\end{cases}
$}} \\
\rowcolor{gray!10} LightThinker \cite{zhang2025lightthinker} & SFT & $\begin{aligned}
\text{Vanilla:Dependency} &= \frac{L_O^2}{2} + L_P \cdot L_O \\
\text{H2O:Dependency} &= \frac{2 L_P L_C + 2 L_O L_C - L_P^2 - L_C^2}{2} \\
\text{LightThinker:Dependency} &= \sum_{t=1}^{L_O} \text{ContextLength}_t
\end{aligned}$ \\
LS-Mixture SFT \cite{yu2025long} & SFT & $\mathcal{L}(D_{\text{long}}) + \mathcal{L}(D_{\text{short}}) = \sum_{(x_i, r_i^L, y_i)} -\log P(r_i^L \oplus y_i \mid x_i) + \sum_{(x_i, r_i^S, y_i)} -\log P(r_i^S \oplus y_i \mid x_i)$ \\
\rowcolor{gray!10} PIR \cite{xiao2025limopro} & SFT & $\text{PIR}_{\theta}(x_i | x_{1:n}) = \log \left( \frac{\text{PPL}(R \setminus \{x_i\})}{\text{PPL}(R)} \right)$ \\
R1-Compress \cite{wang2025r1} & SFT & $\hat{c}_i^* = \arg\max_{\hat{c} \in \tilde{C}_i} \pi_\theta(\hat{c} \mid x, \hat{c}_{<i})$ \\
\rowcolor{gray!10} SF \cite{munkhbat2025self} & SFT & $\text{Relative Length} = \frac{\text{Avg. output length (method)}}{\text{Avg. output length (baseline)}};
\text{Relative Accuracy} = \frac{\text{Accuracy (method)}}{\text{Accuracy (baseline)}}
$
\\
Skip Steps \cite{liu2024can} & SFT & $ M^{\text{standard}}_k = \prod_{(q, a^{(n)}) \in D_k} P(a^{(n)} \mid q)$
\\
\rowcolor{gray!10} SOLAR \cite{li2025solar} & SFT & $J(\theta) = \mathbb{E}_{x \sim D,\, y \sim \pi_\theta(\cdot\,|\,x; t^*, s^*)} \left[ r(y) \right]$
\\
SoftCoT \cite{xu2025softcot} & SFT & $\mathcal{L} = \mathbb{E}_{\mathbf{z} \sim q_{\phi}(\mathbf{z}|\mathbf{x}, \mathbf{y})} \left[ \log p_{\theta}(\mathbf{y}|\mathbf{z}, \mathbf{x}) \right] - D_{\mathrm{KL}} \left( q_{\phi}(\mathbf{z}|\mathbf{x}, \mathbf{y}) \,\|\, p_{\theta}(\mathbf{z}|\mathbf{x}) \right)$
\\
\rowcolor{gray!10} \textsc{L2} \cite{chen2025less} & SFT, Decoding Intervention &
\makecell[c]{
$\hat{z}_t(j) =
\begin{cases}
z_t(j) + \beta, & \text{if } j \in TopK(z_t,k) \text{ and } u_j < \alpha,\\
z_t(j) - \beta, & \text{if } j \in TopK(z_t,k) \text{ and } u_j \ge \alpha,\\
z_t(j), & \text{otherwise.}
\end{cases}$
}
\\
VARR \cite{jang2025varr} & SFT & 
$\begin{array}{c}
\mathcal{L} = - \log p_\theta (y_g, R \mid x) \\
\text{verbosity}(y_g) = \log \frac{p_\theta(y_g \mid R', x)}{p_\theta(y_g \mid R, x)} \\
\text{verbosity}(y_w) - \text{verbosity}(y_g) \le 0
\end{array}$ \\
\rowcolor{gray!10} Token assorted \cite{su2025token} & SFT & $ L(X) = \log p \big(X \,|\, f_{\text{dec}}( q(\bar{X}) ) \,|\, g(P) \big)
+ \sum_{i=1}^{L} \big\| \text{sg}[\bar{X}_i] - q(\bar{X}_i) \big\|_2^2
+ \beta \big\| \bar{X}_i - \text{sg}[q(\bar{X}_i)] \big\|_2^2$
\\
ACPO \cite{cheng2025incentivizing} & SFT, RL & 
\multicolumn{1}{>{\centering\arraybackslash}p{14cm}}{
\makecell[c]{
$R_i = \begin{cases} \max(w_{\text{acc}} R_{\text{acc}} + w_{\text{len}} R_{\text{TLB}} + w_{\text{think}} R_{\text{think}}, 0.1), & y_i \text{ correct} \\ \min(\cdots, -0.1), & y_i \text{ incorrect} \end{cases}$ }}\\
\rowcolor{gray!10} TokenSkip \cite{xia2025tokenskip} & SFT & $\mathcal{L} = \sum_{i=1}^{l} \log P(y_i \mid x, \gamma, y_{<i}; \theta_M)$
\\
VeriThinker \cite{chen2025verithinker} & SVFT & $\min_{\theta} \, \mathbb{E}_{q} \left[ D\left( M_{\theta}(\cdot \mid q), C_i \right) \right]$ \\
\rowcolor{gray!10} CoLaR \cite{tan2025think} & SFT, RL & $\mathcal{L}_{\text{total}} = \mathcal{L}_{\text{comp}} + \mathcal{L}_{\text{latent}}$ \\
CoThink \cite{fan2025cothink} & SFT, RL, Distillation & $\tau(M, D) = \frac{Q_M(D)}{C_M(D)}, \quad \eta(M_R, M_I) = \frac{Q_R C_I}{Q_I C_R}$ \\
\rowcolor{gray!10} Long Short \cite{ning2025not} & SFT, RL & $M(y_i) = \log_2\left(1 + \left( \frac{d_y - d_{\{y_1,\dots,y_i\}}}{d_y} \right) \left( \frac{d_y - d_{y_i}}{d_y} \right) \left( \frac{N_i^{\text{right}}}{N_i^{\text{sum}}} \right) \right) - \delta(y_i)$ \\
\bottomrule[1.1pt]
\end{tabular}}
\end{table*}



\end{document}